\documentclass[lettersize,journal]{IEEEtran}
\usepackage{cite}
\usepackage{amsmath}
\usepackage{amsthm}
\usepackage{amsfonts}
\usepackage{url}
\usepackage{graphicx}
\usepackage{subfigure}
\usepackage{caption}
\usepackage{epstopdf}
\usepackage{multirow}
\usepackage{color,xcolor}
\usepackage{cite}
\usepackage[switch]{lineno}
\usepackage{CJK}
\usepackage{mathrsfs}
\usepackage{algorithm}
\usepackage{algorithmic}
\usepackage{multicol}
\usepackage{mathtools}
\usepackage{booktabs}
\usepackage{amssymb}
\usepackage{hyperref}
\usepackage{comment}
\usepackage{makecell}
\usepackage{bbding}

\def\M{\mathcal{M} }

\def\F{\mathbf{F}}
\def\T{\mathbf{T}}
\def\V{\mathbf{V}}

\def\Dset{\mathcal{D} }

\def\W{\mathcal{W} }
\def\L{\mathcal{L} }
\newcommand{\R}{\ensuremath{\mathbb{R}}}

\newif\ifannotated

\annotatedtrue

\ifannotated

\newcommand{\delete}[1]{{\color{red}{\sout{#1}}}}

\newcommand{\margincomment}[1]{\marginpar{$\Rightarrow$\color{red}\fbox{\parbox{\linewidth}{\color{black}\scriptsize#1}}}}
\else

\newcommand{\delete}[1]{{\ignorespaces}}

\newcommand{\margincomment}[1]{}
\fi

\begin{document}

\title{GeodesicPSIM: Predicting the Quality of Static Mesh with Texture Map via Geodesic Patch Similarity}
\author{Qi Yang$^*$,
        Joel Jung,
        Xiaozhong Xu,~\IEEEmembership{Member,~IEEE},
        and Shan Liu, ~\IEEEmembership{Fellow,~IEEE}
        \IEEEcompsocitemizethanks{\IEEEcompsocthanksitem Q. Yang, J. Jung, X. Xu, and S. Liu are from Tencent Media Lab, (e-mail: chinoyang@tencent.com, \{joeljung, xiaozhongxu, shanl\}@global.tencent.com)}
         \thanks{*: Corresponding author: Q. Yang}
}

\IEEEtitleabstractindextext{
\begin{abstract}
Static meshes with texture maps have attracted considerable attention in both industrial manufacturing and academic research, leading to an urgent requirement for effective and robust objective quality evaluation. However, current model-based static mesh quality metrics (i.e., metrics that directly use the raw data of the static mesh to extract features and predict the quality) have obvious limitations: most of them only consider geometry information, while color information is ignored, and they have strict constraints for the meshes' geometrical topology. Other metrics, such as image-based and point-based metrics, are easily influenced by the prepossessing algorithms, e.g., projection and sampling, hampering their ability to perform at their best. In this paper, we propose Geodesic Patch Similarity (GeodesicPSIM), a novel model-based metric to accurately predict human perception quality for static meshes. After selecting a group keypoints, 1-hop geodesic patches are constructed based on both the reference and distorted meshes cleaned by an effective mesh cleaning algorithm. A two-step patch cropping algorithm and a patch texture mapping module refine the size of 1-hop geodesic patches and build the relationship between the mesh geometry and color information, resulting in the generation of 1-hop textured geodesic patches. Three types of features are extracted to quantify the distortion: patch color smoothness, patch discrete mean curvature, and patch pixel color average and variance. To the best of our knowledge, GeodesicPSIM is the first model-based metric especially designed for static meshes with texture maps. GeodesicPSIM provides state-of-the-art performance in comparison with image-based, point-based, and video-based metrics on a newly created and challenging database. We also prove the robustness of GeodesicPSIM by introducing different settings of hyperparameters. Ablation studies also exhibit the effectiveness of three proposed features and the patch cropping algorithm. The code is available at \url{https://multimedia.tencent.com/resources/GeodesicPSIM}.

\end{abstract}

}

\maketitle

\IEEEdisplaynontitleabstractindextext

%
\IEEEpeerreviewmaketitle

\section{Introduction}\label{sec:intro}

\begin{figure}[t]
	\centering
    \includegraphics[width=1\linewidth]{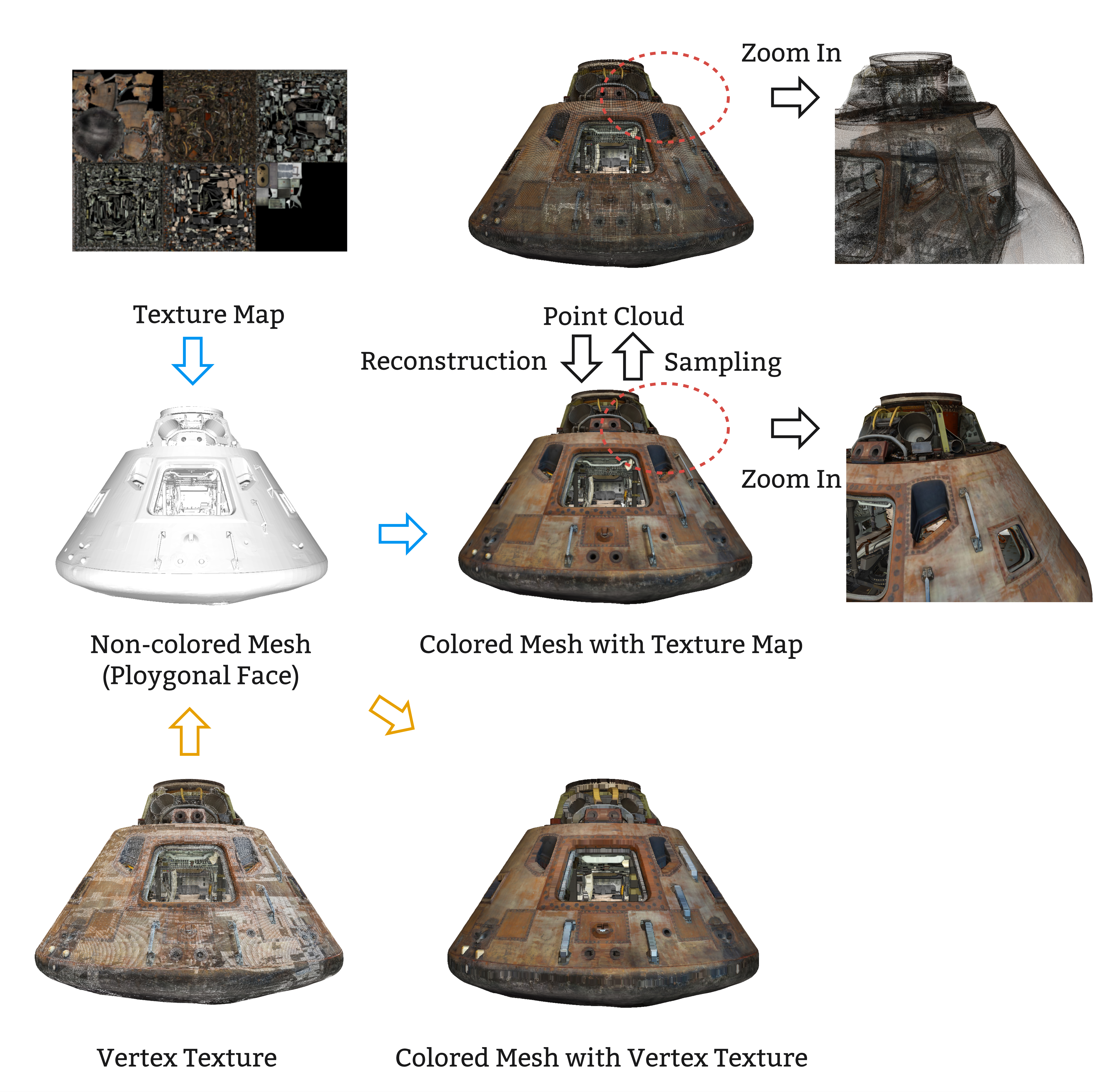}%
	\caption{Illustration of 3D mesh. The samples are from the Tencent Static Mesh Dataset (TSMD)\cite{yang2023tsmd}.}
	\label{fig:meshVSpc}
\end{figure}

A 3D mesh is a collection of vertices in the 3D space, connected through edges that compose polygonal faces. With additional attributes, such as normal vectors and texture coordinates {(e.g., UV coordinates)}, static meshes are widely used in many areas, such as gaming, animation, medical imaging, and industrial manufacturing. The 3D mesh and the 3D point cloud, which is another type of prevalent 3D media consisting of scattered points, are mutually convertible as illustrated in Fig. \ref{fig:meshVSpc}. The differences are that 3D meshes have a regular topology and a continuous surface, which can be transformed geometrically and textured, leading to a flexible and efficient rendering. Due to their unique characteristics, static meshes attract considerable attention in modern industrial design, such as rapid prototyping, digital fabrication, and precision measurement, and also become a hotspot for academic research.

Static meshes can be divided into three types regarding the texture form: non-colored mesh, colored mesh with vertex texture, and colored mesh with texture map,as shown in Fig. \ref{fig:meshVSpc}. {For non-colored mesh, only face and vertex information is provided. For colored mesh with vertex texture, it additionally provides RGB information for each vertex (similar to colored point cloud) based on non-colored mesh, the color inside each face is interpolated based on polygon vertices. For colored mesh with texture map, the color information is provided by an individual texture map, a group of texture coordinates (i.e., UV) are used corresponding to each face, indicating the detailed texture of each face via a certain area in the texture map. Generally, for a mesh with complex texture, colored mesh with texture map can use fewer faces compared to colored mesh with vertex texture.}
In this paper, we focus on static meshes with texture maps due to their common wide utilization in academic and industrial fields. 

\begin{figure}[h]
	\centering
    \includegraphics[width=0.6\linewidth]{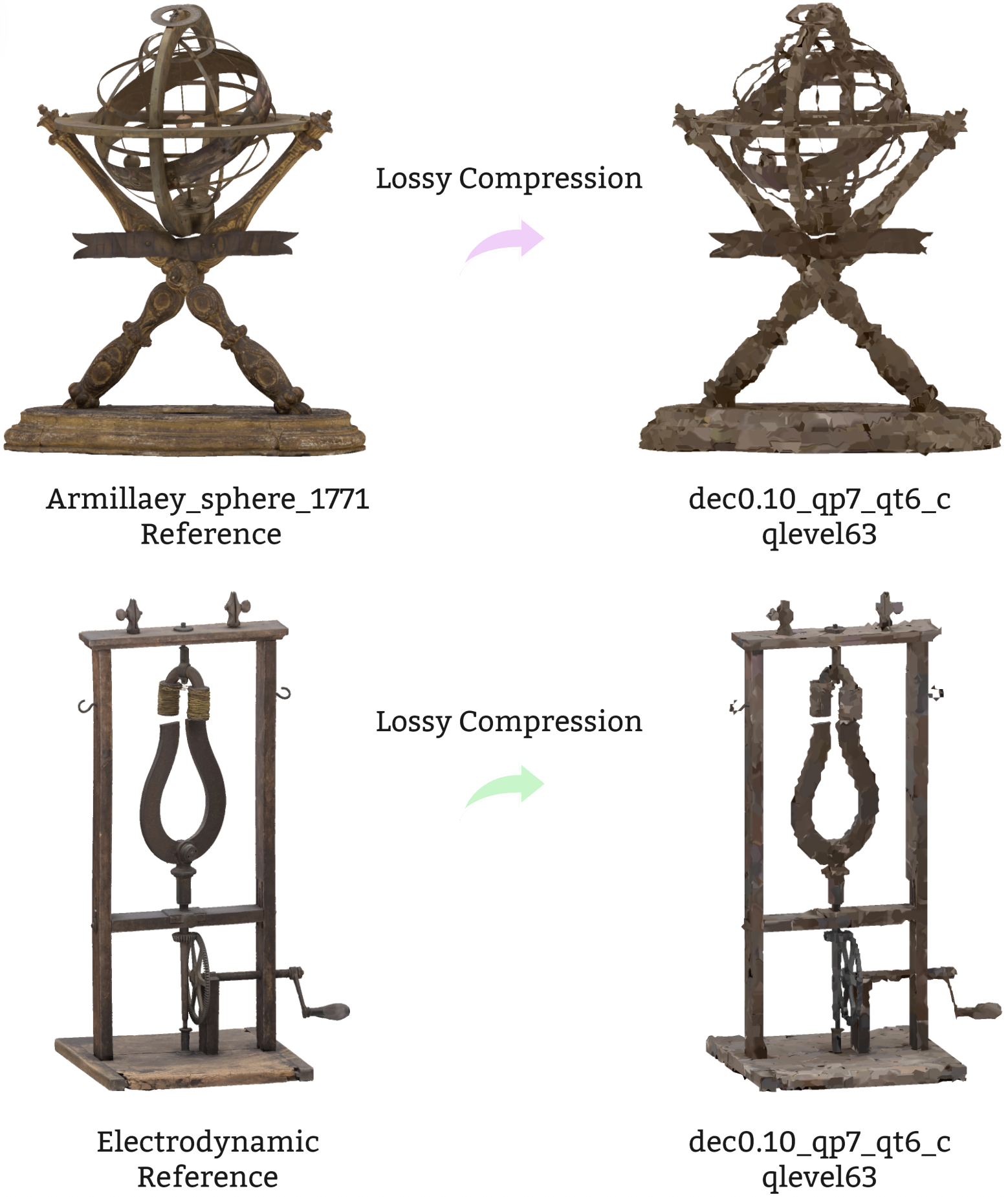}%
	\caption{Snapshots of lossy compressed meshes. "dec", "qp", "qt", and "cqlevel" are compression parameters defined in section \ref{sec:TSMD}.}
	\label{fig:dis-illustration}
\end{figure}

Given the substantial volume of static mesh data, static mesh compression emerges as an extremely important and unavoidable technology \cite{VVM-cfp}, raising new requirements for the quality evaluation of static meshes in terms of lossy compression \cite{tip-2-coding, tip-4-coding}, as illustrated in Fig. \ref{fig:dis-illustration}. Rate-distortion (RD) curves are used to measure compression efficiency \cite{RD-peng2005technologies}, which means a reliable quality evaluation can help to select a superior codec tool, select an optimal compression configuration, and save bandwidth and other resource costs.

Undoubtedly, the most reliable quality evaluation method is to conduct subjective experiments, collecting enough subjective scores to calculate mean opinion scores (MOS). However, the subjective experiment is expensive, time-consuming, and requires a rigorous testing environment, leading to extreme inconvenience in practical applications. Therefore, an effective and robust objective metric is needed to replace subjective experiments and facilitate the study of mesh compression. 

Objective quality assessment of static meshes has made significant progress since a series of model-based metrics (i.e., metrics that directly use the raw data of the static mesh to extract features and predict the quality) 
 are suggested to quantify the distortion of static mesh geometry by introducing structure features, such as curvature \cite{MSDM-lavoue2006perceptually}, smoothness \cite{DWPM-corsini2007watermarked, GL-karni2000spectral}, dihedral angles \cite{dame-vavsa2012dihedral}, etc., showing better correlations with human perception than metrics based on point-wise features \cite{rms-cignoni1998metro}. However, these earlier model-based metrics are not sufficiently robust to handle the complexities involved in current static mesh quality evaluation tasks for the following reasons: first, these metrics only consider geometry features, targeting the non-colored static meshes, while currently, the textured static meshes are the spotlight due to broader application scenarios; second, they have strict constraints regarding the meshes' geometrical topology, requiring both the reference and the distorted mesh have perfect manifold surface and share the same vertex density, the same connectivity, or the same level of details. Therefore, the appearance of new demands has spurred research into objective static mesh quality assessment. In this paper, we focus on the full-reference metric, in which the reference mesh is provided as evidence to assist in the quality prediction of distorted meshes.   

Given the rapid development in the fields of image and point cloud quality assessment, WG7 - MPEG Coding of 3D Graphics and Haptics proposed two strategies to measure static mesh quality, i.e., image-based and point-based metrics \cite{MPEG-MESH-metric}. Some research also suggested using rendered mesh videos to infer mesh quality \cite{nehme2020visual}.  These methods all need to convert the static meshes into other media formats, such as image, point cloud, and video, by some preprocessing algorithms like projection and sampling, which unavoidably introduce errors and hamper the metrics' ability to perform at their best. For instance, projected images may contain background information that can mask the influence of distortion and reduce metric performance, and unavoidable errors occur when an improper projection view is selected, preventing image-based metrics from making accurate predictions. In addition, preprocessing algorithms require extra computational resource overhead, such as rendering of static mesh needs extra time, and sampling methods typically operate on faces in a per-face manner. 

In light of the above analysis and explanation, it can be concluded that an effective and robust model-based metric, targeting static meshes with texture maps that undergo different types of distortion, needs to be designed. {Two key problems need to be considered for design effective full-reference model-based metrics. First, either reference or distorted mesh may have redundant and invalid information, and current available model-based metrics cannot extract features for this kind of mesh and limited the real application. Second, current model-based metrics are mostly based on geometry features and ignore texture distortion. Based on the experience of quality assessment, designing effective color feature extraction methods is the guarantee of accurate quality prediction results.}

 {{In this paper, we propose an effective model-based metric, called Geodesic Patch Similarity (GeodesicPSIM), inspired by multidisciplinary research such as neuroscience, graph signal processing (GSP), and conformal geometry. GeodesicPSIM is the first model-based metric especially designed for static mesh with texture map, solving the key problems mentioned above by thoughtfully designed effective algorithms. First, a mesh cleaning algorithm is proposed to solve the first key problem, which can detect and remove redundant and invalid information such as duplicated vertices, null faces, etc. Second, a cluster of keypoints is selected, centered on which a group of 1-hop geodesic patches is constructed as feature extraction units.   Third, a two-step patch cropping algorithm is proposed to resize 1-hop geodesic patches to a suitable size. Subsequently, a patch texture mapping module is proposed to generate 1-hop textured geodesic patches. Effective pixels {(i.e., pixels have definite x-y coordinates on the texture map)}, derived from the texture map, are extracted for each face contained in 1-hop geodesic patches. Finally, three types of features, i.e., patch color smoothness, patch discrete mean curvature, and patch pixel color average and variance, are proposed to quantify the influence of mesh distortion, addressing the second key problem. Patch color smoothness and discrete mean curvature are inspired by the observations in neuroscience that the human visual system is sensitive to high-frequency distortion \cite{yang2020inferring,sal-ev-pami} and processes color and form information separately \cite{rentzeperis2014distributed}. Patch pixel color average and variance are proposed to quantify the distortion within each face, in which the effective pixels of each face are extracted from the texture map. GeodesicPSIM scores are calculated by combining all feature similarities with a simple method.}}

To prove the superiority of GeodesicPSIM, a newly created and challenging static mesh quality assessment database, called TSMD, is introduced. Then GeodesicPSIM and 16 other state-of-the-art (SOTA) metrics are evaluated on TSMD. Three correlation indicators are reported, including Pearson's linear correlation coefficient (PLCC), Spearman's rank order correlation coefficient (SRCC), and root mean square error (RMSE). The experimental results suggest the reliable and superior performance of GeodesicPSIM in MOS prediction for TSMD with PLCC, SRCC, and RMSE at 0.81, 0.81 and 0.69, which is the first in all metrics. The robustness of GeodesicPSIM is also examined by adjusting the selection of hyperparameters, revealing that GeodesicPSIM presents stable and predictable performance variations in various scenarios. Ablation studies demonstrate the importance and necessity of feature pooling and patch cropping. The main contribution of this paper is summarized as follows:

\begin{itemize}
	\item to our knowledge, the proposed GeodesicPSIM is the first model-based metric especially designed for static mesh with texture map. We solve the key problems of quality assessment for static mesh with texture map and propose three types of features for robust, effective quality prediction.
 \item we test GeodesicPSIM using a newly created and challenging database, exhibiting the SOTA results in MOS prediction.
 \item we perform extensive experiments to show the robustness and generalizations of GeodesicPSIM.
\end{itemize}

The remainder of this paper is as follows. Section \ref{sec:RelatedWork} presents the related work about mesh objective quality metrics. Section \ref{sec:TSMD} introduces the database used to evaluate objective metrics, reporting the weaknesses of the SOTA metrics. Section \ref{sec:metric design} details the design of GeodesicPSIM. Section \ref{sec:exp} illustrates the experimental results and analyzes the robustness and generalization of GeodesicPSIM. Section \ref{sec:conclusion} concludes the article and highlights future work.

\section{Related work}\label{sec:RelatedWork}
\subsection{Mesh Metric}
Objective quality evaluation for static mesh has been studied for decades, it can be roughly divided into four types: image-based, point-based, video-based, and model-based metrics. 

Image-based, point-based, and video-based metrics all require preprocessing algorithms before quality prediction, that is, converting static meshes into other media formats like images, processed video sequences (PVSs), and point clouds, and then using mature and well-accepted metrics with their respective fields to infer quality. Image-based metrics use multiple projected images from static meshes with different viewpoints as evidence serving the metrics such as $\rm geo_{psnr}$ \cite{MPEG-MESH-metric}, $\rm rgb_{psnr}$, etc. More recently, \cite{nehme2022textured} proposed a learning-based image-based metric for mesh specifically. Point-based metrics first sample the static meshes via classical sampling algorithms, such as grid sampling, face sampling, surface subdivision sampling, etc., then the SOTA point cloud objective metrics are involved to predict quality scores: point-to-point (D1) \cite{D1-Mekuria2016Evaluation}, point-to-plan (D2) \cite{tian2017geometric}, $\rm PCQM$ \cite{meynet2020pcqm}, MS-GraphSIM \cite{ms-zhang2021ms}, TCDM\cite{zhangTCDM}, MM-PCQA\cite{zhang2022mm}, SGR \cite{pcqa_new2}, and DHCN \cite{pcqa_new1}, etc. Video-based metrics first define a smooth camera path to capture continuous images to generate PVS, then image/video quality assessment (IQA/VQA) metrics \cite{min2021screen, min2024perceptual, zhai2020perceptual} are utilized to infer quality, such as SSIM \cite{wang2004ssim}, MSSIM \cite{wang2003multiscale}, and VMAF \cite{vmaf-li2016toward}. However, the preprocessing algorithms required by the above metrics may incur unexpected bias, hampering the metrics' ability to show their best performance.

Unlike the above-mentioned three methods, model-based metrics do not need media type conversion, they directly use the raw data of the static mesh to extract features and predict the quality. From the earliest point-to-mesh  \cite{rms-cignoni1998metro}, based on point-wise features, to metrics using structure features such as MSDM \cite{MSDM-lavoue2006perceptually}, DWPM \cite{DWPM-corsini2007watermarked}, and GL \cite{GL-karni2000spectral}, model-based metrics gradually show better performance by taking into account more and more characteristics of the human visual system. For example, MSDM \cite{MSDM-lavoue2006perceptually} introduced a surface curvature to quantify mesh quality. DWPM \cite{DWPM-corsini2007watermarked} made use of global roughness to measure the distortion of the mesh surface. DAME \cite{dame-vavsa2012dihedral} detected variations in the structure of the mesh geometry based on the estimation of dihedral angles. GL \cite{GL-karni2000spectral} proposed another method for calculating surface smoothness using the geometric Laplacian of vertices. \cite{zhang2022no} proposed a no-reference mesh metric based on 3D natural scene statistics.

However, the model-based metrics cited above only consider geometry distortions while texture distortions are ignored. Consequently, it leads to an inadequate quality prediction considering texture information that might mask the characteristics of geometry \cite{TMM-masking,yangTMM}. In addition, they also have strict constraints on distorted meshes, such as sharing the same connectivity, the same vertex density, or the same level of detail \cite{MPEG-MESH-metric}. Duplicated vertices, duplicated faces, and null faces can even affect the normal operation of the metrics, indicating that they have poor tolerance for invalid and interference information, which limits the metrics' performance in real applications. Therefore, a new effective and robust metric is needed to facilitate the study of mesh compression and other visual quality-related tasks.
\subsection{Mesh Dataset}
Dataset is the first step to study mesh quality. At the early stage, mesh datasets are all non-colored meshes. For example, ~\cite{christaki2019subjective} collected eight meshes: four 3D human reconstructions and four scanned objects as references. Three different codecs are applied to generate distorted samples. The subjective experiment was conducted using a VR application. ~\cite{DWPM-corsini2007watermarked} proposed a dataset with 2D monitors as viewing equipment, four mesh objects with watermarking distortions are generated to collect subjective scores. 

More recently, with the development of mesh generation technologies, more datasets have been proposed with respect to texture meshes. ~\cite{nehme2020visual} constructed a colored mesh with vertex texture, using a VR environment to score meshes with four types of distortions.  \cite{guo2016subjective} released the first dataset of colored mesh with texture map, consisting of five references and five types of distortion.  ~\cite{nehme2022textured} proposes 55 references with a crowdsourcing environment to collect subjective scores. Besides, it used pseudo MOSs to label around 300K samples. \cite{zhang2023advancing} published a dataset called SJTU-H3D, which comprises 40 high-quality reference digital humans and 1,120 labeled distorted counterparts.  

Considering the development of mesh compress tool, a new dataset called \cite{yang2023tsmd} is constructed referring to AOMedia VVM texture mesh compression anchor \cite{VVM-cfp}. This dataset can be regarded as the latest benchmark for the mesh quality assessment study, which will be introduced in detail in Section \ref{sec:TSMD}.

\section{TSMD-A challenging static mesh quality assessment database}\label{sec:TSMD}

In this section, a newly created database with rich content, reliable MOSs, and useful supplementary materials (PVSs and bitrate), called TSMD, is introduced \cite{yang2023tsmd} \footnote{\url{https://multimedia.tencent.com/resources/tsmd.}}. Due to the diversity of distorted meshes, the SOTA metrics only achieve a correlation of around 0.75, serving as a catalyst for the GeodesicPSIM study. 

\subsection{Database Creation}
\subsubsection{Content description}
TSMD compiled 42 publicly available meshes from reputable sources, which encompass a rich category of static meshes, such as human characters, animals, plants, buildings, indoor scenes, inanimate objects, etc. The essential information of the database is summarized in Table \ref{tab:TSDM}

\begin{table}[!ht]
    \centering
    \caption{Summary information of the TSMD.}
    \begin{tabular}{|c|c|}
    \hline
        Number of meshes & 42 \\ \hline
        \thead{Distortions applied to the mesh\\ (successively)} & \thead{Dequantization, Triangulation, \\ Decimation, Draco compression} \\ \hline
        \thead{Distortions applied to the texture \\(successively)} & \thead{Downscaling, \\ AV1 (libaom) compression} \\ \hline
        Levels of distortion & 5 \\ \hline
        Total number of distorted meshes & 210 \\ \hline
        \thead{Total number of scores \\collected via crowdsourcing} &10320\\ \hline
         \thead{Remaining scores after \\ outlier removal} &  9468 \\ \hline
    \end{tabular}
    \label{tab:TSDM}
\end{table}

\subsubsection{Distortion generation}
\begin{figure}[h]
	\centering
    \includegraphics[width=0.6\linewidth]{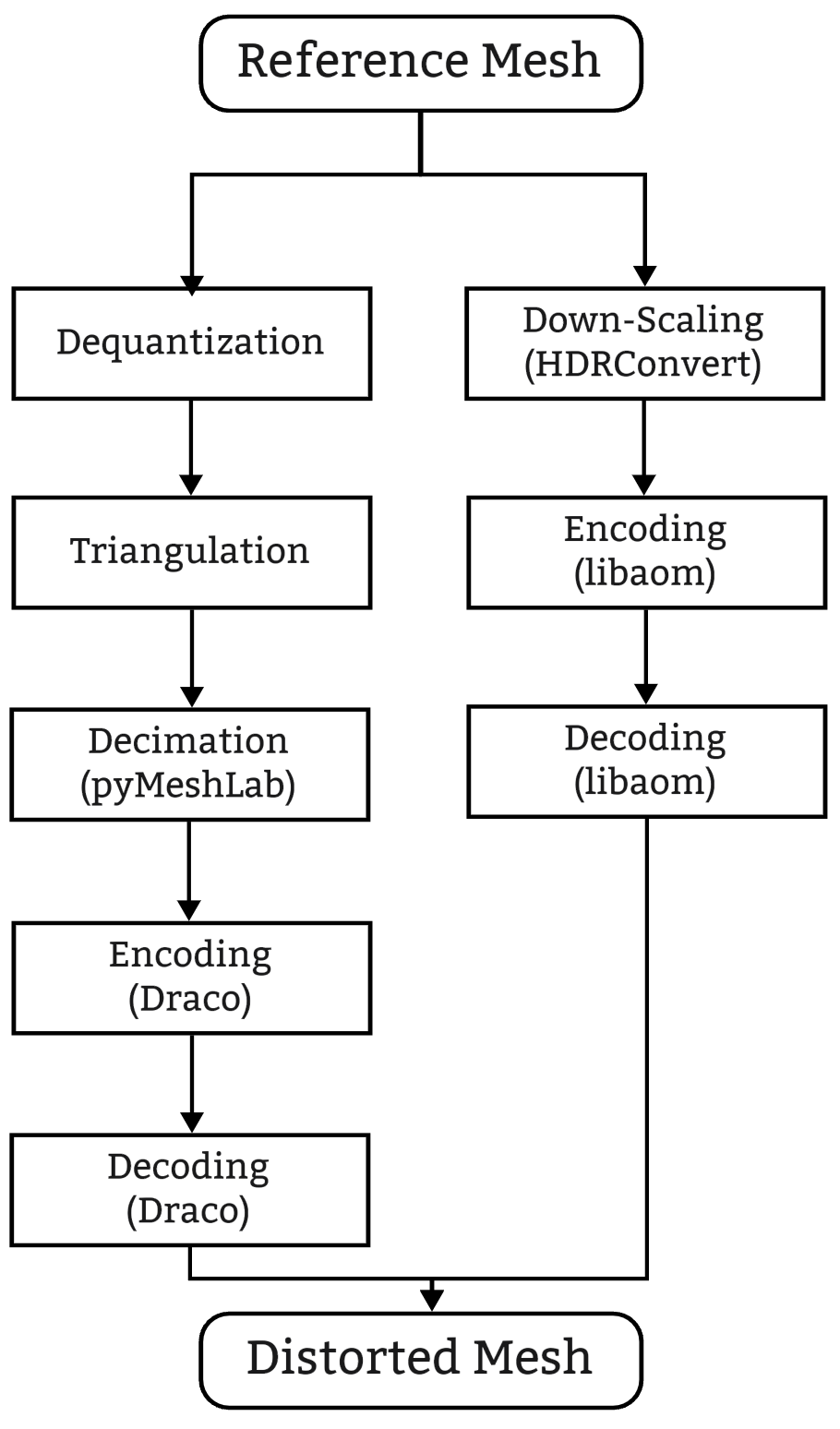}%
	\caption{Static mesh compression anchor.}
	\label{fig:vvmanchor}
\end{figure}

Four distinct types of distortion are successively introduced into the reference meshes, corresponding to the distortion applied in the context of the VVM polygonal static mesh coding Call for Proposals (CfP) of AOMedia \cite{VVM-cfp} depicted in Fig. \ref{fig:vvmanchor}, and generate 210 distorted meshes.  

Each mesh successively undergoes dequantization, triangulation, decimation, encoding, and decoding with Draco. Each texture map is downscaled to meet the following criteria: the vertical/horizontal ratio is maintained, and neither the horizontal nor the vertical resolution exceeds 4096 pixels. Subsequently, AV1 compression is applied. For each distortion, five encoding configurations are chosen, consisting of the following parameters: decimation level (dec), Draco quantization parameter for position attribute (qp) and for the texture attribute (qt) (normals are not quantized), and AV1 quantization (cqlevel). This allows us to achieve five RD points, evenly spread across practical quality ranges that are applicable to various use cases.

\subsubsection{PVS generation}
Processed Video Sequences (PVSs) are generated from both source and distorted content using a conventional approach. These 2D videos depict camera paths that simulate typical user motions and are displayed during subjective tests to enable the evaluation process. The open3D library is used to render the content from distorted meshes and textures. The object is centered, and its size is adjusted to occupy a significant portion of the frame. A rotational movement is applied to the object, following a cosine function that gradually decreases the rotation speed until near stop and then increases it again. For two meshes that represent scenes inside a building, the camera is positioned inside the scene before the rotational movement is applied. The resulting PVSs are created with a frame rate of 30 fps and a duration of 18 seconds and are encoded using the FFMPEG library, with the x264 encoder and a constant rate factor equal to 10 to ensure visually lossless encoding. This approach is in line with the guidelines proposed in \cite{Joel-m57896}.

\subsubsection{Subjective experiment}
A crowdsourcing methodology is employed to conduct subjective tests, which involves a proprietary interface for downloading and streaming the PVS to participants. Before the rating session begins, instructions are given to ensure the clarity and consistency of the test. Additionally, a training session is conducted to replicate the rating session, where participants are asked to assign scores within the range of "possible" scores for 8 PVS, excluding any unrealistic scores. This serves as a qualification criterion for the actual rating session. The methodology follows the ITU-T P.910 Recommendation \cite{11scale-rating} and is adapted to the crowdsourcing approach. A double-stimulus impairment scale is used, where the source content and the distorted content are shown in pairs, in random order, before the score is requested, using a five-level scale to rate the impairments. The participants are "naïve" viewers, not familiar with research related to meshes, and consist of 74 students aged 17 to 25 years.


\subsubsection{Outlier detection}
In the realm of crowdsourcing, the detection and elimination of outliers is of the utmost importance. To tackle this problem, certain "trapping" PVS are deliberately included in the rating sessions. These PVS can be divided into two categories: 1- extremely low-quality PVS, for which extremely low scores are expected, and 2- duplicated PVS, for which close scores are anticipated. By incorporating these trapping PVS, it is possible to identify and filter out any outliers during the evaluation process. In addition, the correlation between the average score (calculated based on all available subjective scores for each PVS from multiple viewers) and each group of raw data (subjective scores from one viewer) is calculated. If the correlation value is below 0.8, the corresponding data are considered unreliable and are consequently excluded from the database. As a result, 852 scores were removed from a total of 10320. It is ensured that each PVS has a remaining number of scores above 15 to compute the Mean Opinion Score (MOS), once outliers are eliminated.


\subsection{Database validation}
The diversity of content and the accuracy of MOS are hereby comprehensively discussed and summarized to facilitate the presentation of key information.

\begin{figure*}[h]
	\centering
    \includegraphics[width=1\linewidth]{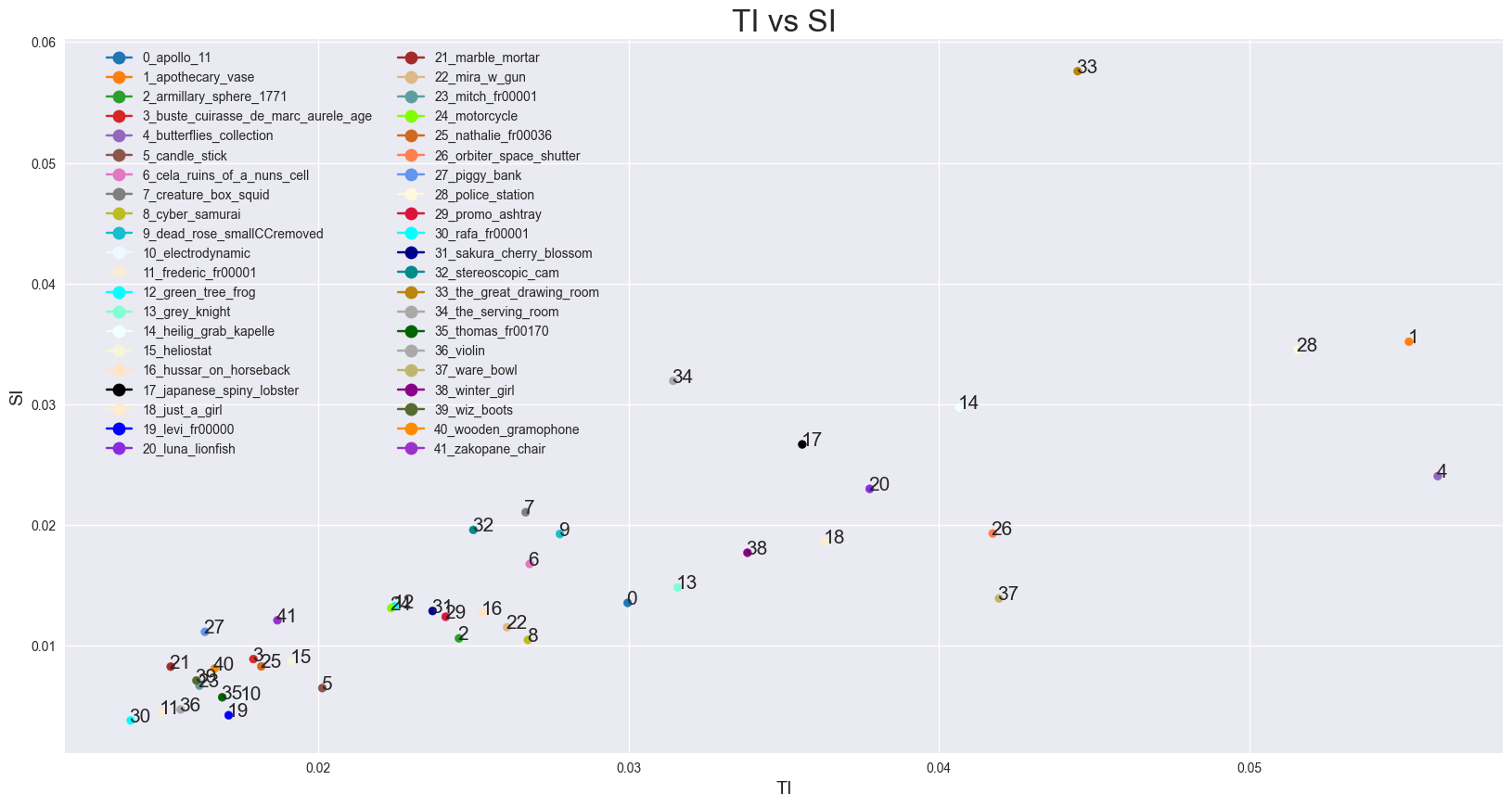}%
	\caption{SI vs TI for TSMD.}
	\label{fig:siti}
\end{figure*}

 First, to measure the diversity of the database, spatial information (SI) and temporal information (TI) are calculated based on PVS and are illustrated in Fig. \ref{fig:siti}. Both indicators reveal the difference in geometry and texture characteristics of 3D meshes under different viewpoints. The indoor scene meshes (e.g., the\_great\_drawing\_room), enrich the diversity of content in terms of spatial information significantly.

\begin{figure}[h]
	\centering
    \includegraphics[width=0.50\linewidth]{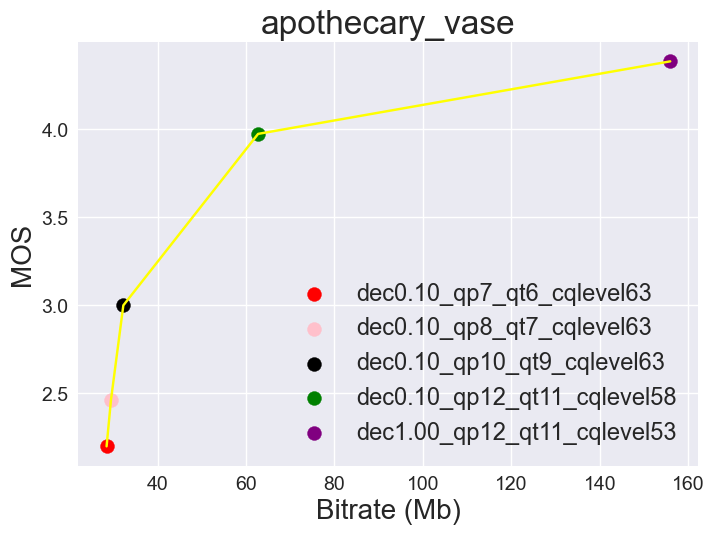}%
    \includegraphics[width=0.50\linewidth]{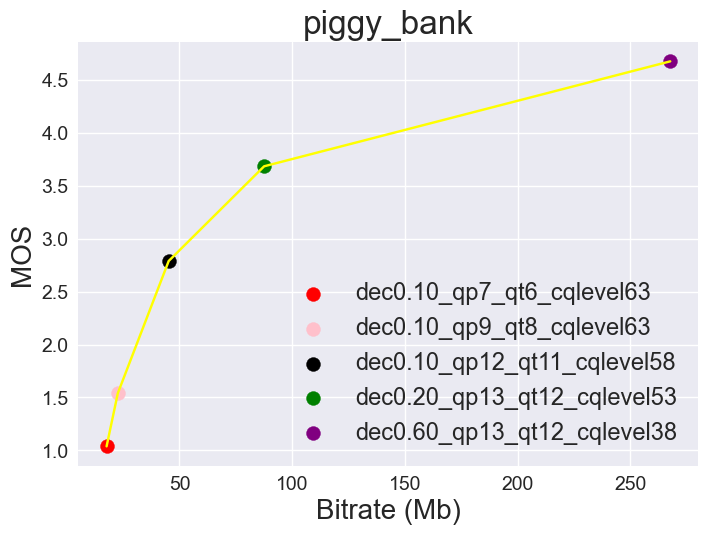}%
	\caption{Example of monotonic bitrate vs MOS curves.}
	\label{fig:bivsmos}
\end{figure}
\begin{figure}[h]
	\centering
    \includegraphics[width=1\linewidth]{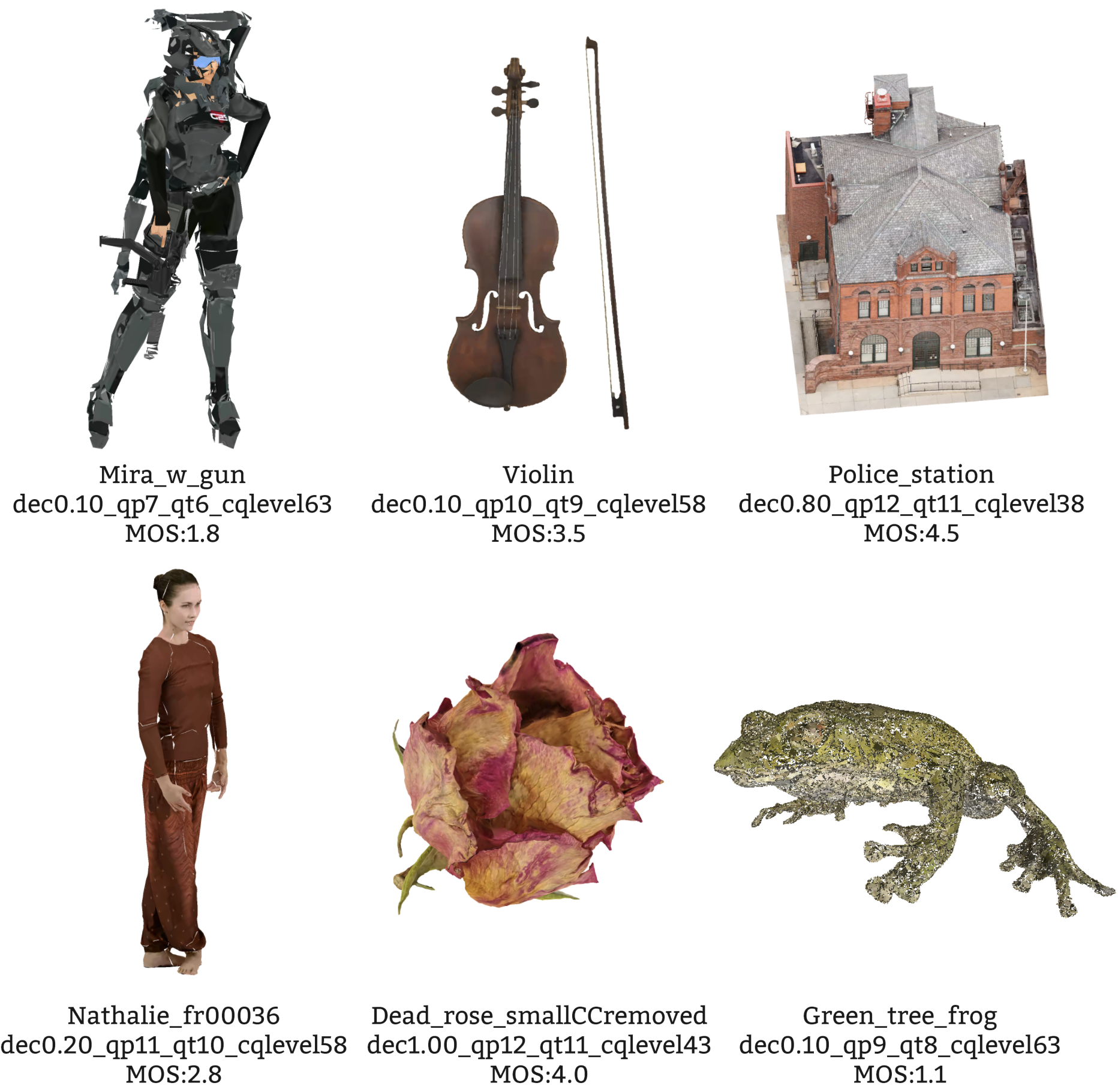}%
	\caption{Illustration of meshes with MOSs.}
	\label{fig:MOSexam}
\end{figure}

 {{Second, to verify the MOS accuracy of the database, the correlations between MOS and bitrate are carefully examined. Most bitrate vs MOS curves show perfect monotony as shown in Fig. \ref{fig:bivsmos}, satisfying the intuition that a higher bitrate that provides more delicate visual information should have a higher MOS. Some meshes with MOS values are illustrated in Fig. \ref{fig:MOSexam}. }}

\subsection{SOTA metric performance on TSMD}

The results of eight SOTA objective metrics, that is, two image-based metrics, $\rm geo_{psnr}$ \cite{MPEG-MESH-metric} and $\rm yuv_{psnr}$ \cite{MPEG-MESH-metric}, three point-based metrics, D1 \cite{D1-Mekuria2016Evaluation}, D2 \cite{tian2017geometric}, and $\rm PCQM_{psnr}$ \cite{meynet2020pcqm}\cite{MPEG-MESH-metric}, three video-based metrics, PSNR, SSIM \cite{wang2004ssim}, and VMAF \cite{vmaf-li2016toward}, are reported in Table \ref{tab:vcip_result}, in which the best is $\rm PCQM_{psnr}$ \cite{meynet2020pcqm} achieving a correlation around 0.75.


\begin{table}[!ht]
    \centering
    \caption{Metric Performance on TSMD}
    \begin{tabular}{|c|c|c|c|}
    \hline
        Metric & PLCC & SRCC & RMSE \\ \hline
        $\rm geo_{psnr}$  & 0.73  & 0.73  & 0.80  \\ \hline
        $\rm yuv_{psnr}$  & 0.68  & 0.68  & 0.85  \\ \hline
        D1  & 0.72  & 0.65  & 0.80  \\ \hline
        D2  & 0.54  & 0.47  & 0.98  \\ \hline
        $\rm PCQM_{psnr}$  & \textbf{0.76}  & \textbf{0.75}  & \textbf{0.76}  \\ \hline
        PSNR  & 0.53  & 0.54  & 0.99  \\ \hline
        SSIM  & 0.51  & 0.67  & 0.85  \\ \hline
        VMAF  & 0.70  & 0.51  & 1.00  \\ \hline
    \end{tabular}
    \label{tab:vcip_result}
\end{table}

\subsection{Summary}
TSMD provides diverse content and reliable MOS. However, the best metrics are shown to only achieve a correlation of around 0.75 because of the diverse content of TSMD. Previous research often reports high PLCC and SRCC correlations (around 0.9) for the top-performing metrics, indicating that TSMD is a challenging database and motivating the design of new metrics.


\section{GeodesicPSIM: a model-based metric for mesh quality assessment}\label{sec:metric design}

\begin{figure*}[h]
	\centering
    \includegraphics[width=1.0\linewidth]{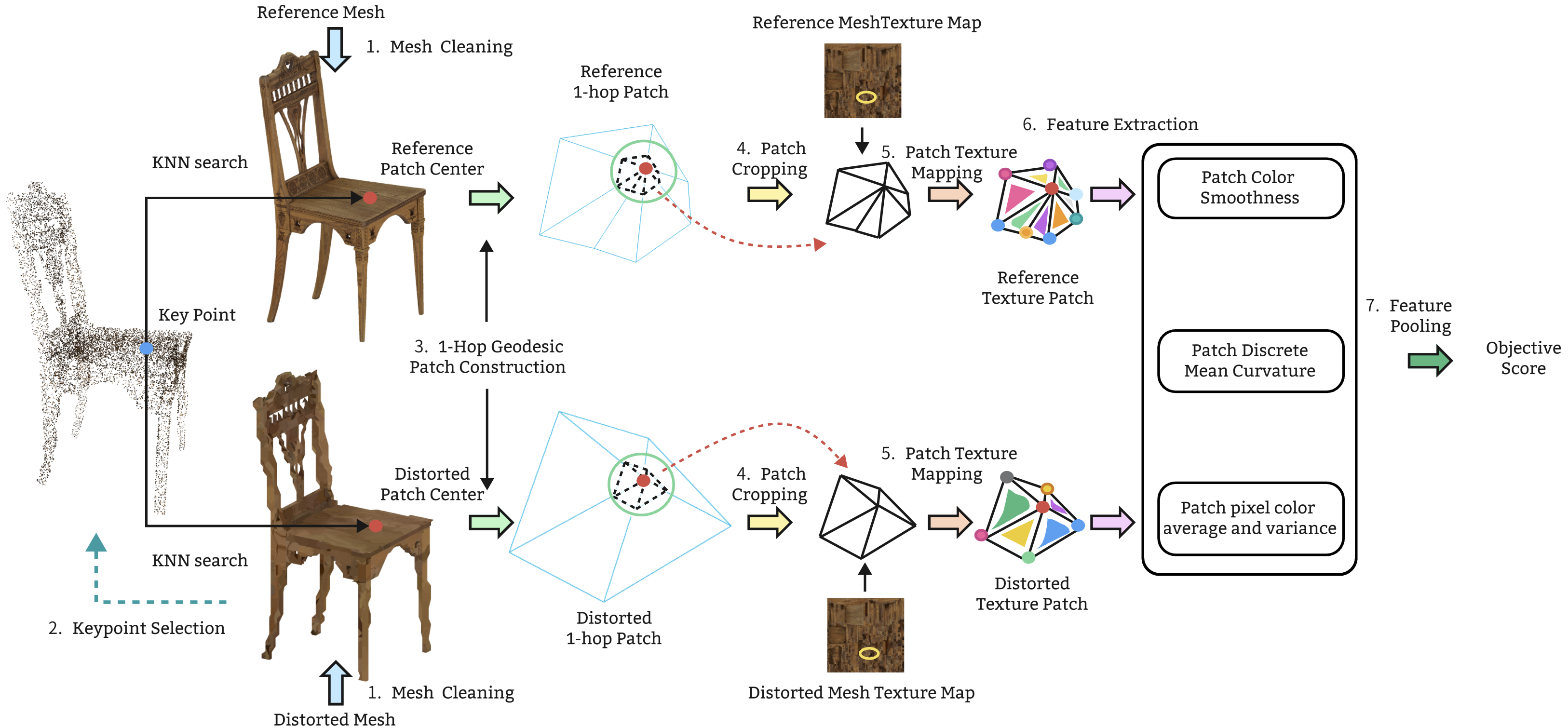}%
	\caption{Flowchart of GeodesicPSIM.}
	\label{fig:flowchart}
\end{figure*}

In this section, the construction of GeodesicPSIM is introduced. It consists of seven steps: mesh cleaning, keypoint selection, 1-hop geodesic patch construction, patch cropping, patch texture mapping, feature extraction, and feature pooling. Fig. \ref{fig:flowchart} illustrates the GeodesicPSIM flowchart.

\subsection{Mesh cleaning}
Different from traditional 2D media, such as images and videos, both reference and distorted meshes are often accompanied by redundant and invalid information, which is generally visually useless but increases computation complexity of metric and even interferes with feature extraction, such as duplicated vertices, null faces, etc.  To ensure the stability of the metric, we propose a mesh cleaning algorithm based on visual fidelity as the first step of GeodesicPSIM. First, duplicated and unreferenced vertices are detected and removed from the vertex matrix. Second, the vertex indexes are updated for the face matrix, after which duplicate and null faces are detected and removed from the face matrix. Based on the new vertex and face matrix, the texture matrix is updated. Repeat the above two steps until no duplicated or unreferenced vertices are detected.

For better understanding, the subscripts $r$ and $d$ are used to indicate the content of reference and distorted meshes in the following sections, while no subscript means the operand/processing is general for both the reference and distorted meshes/objects.

Mathematically, for a mesh, $ \hat{\mathbf{M}}$, the vertex matrix, face matrix, texture matrix are defined as $\hat{\M}_{\V}\in \R^{v^{'}\times3}$, $\hat{\M}_{\F}\in \R^{f^{'}\times3}$, $\hat{\M}_{\T}\in \R^{v\times2}$, the cleaned mesh, $ {\mathbf{M}}$, can be generated using \textbf{Algorithm \ref{al:cleaning}}. 

\begin{algorithm}[h]
\caption{Mesh cleaning }
\label{al:cleaning}
\begin{algorithmic}[1]
    \REQUIRE The raw mesh, $ \hat{\mathbf{M}}$, including vertex matrix $\hat{\M}_{\V}\in \R^{v^{'}\times3}$, face matrix $\hat{\M}_{\F}\in \R^{f^{'}\times3}$, and texture matrix $\hat{M}_{\T}\in \R^{v^{'}\times2}$
    \ENSURE the cleaned mesh, $ {\mathbf{M}}$, including vertex matrix $\M_{\V}\in \R^{v\times3}$, face matrix $\M_{\F}\in \R^{f\times3}$, and texture matrix $\M_{\T}\in \R^{v\times2}$
    \STATE Detecting duplicated vertex $\mathbf{Dv}$ and unreferenced vertex $\mathbf{Uv}$ 
    \WHILE{$\mathbf{Dv}\cup\mathbf{Uv}\neq \emptyset$}
    \STATE Removing the duplicated vertex and unreferenced vertex
    \STATE Updating the vertex index in face matrix $\hat{\M}_{\F}$, updating the texture matrix $\hat{\M}_{\T}$ corresponding to vertex matrix
    \STATE Detecting duplicated face $\mathbf{Df}$ and null face $\mathbf{Nf}$
    \IF{$\mathbf{Df}\cup\mathbf{Nf}\neq \emptyset$}
    \STATE Removing duplicated face and null face
    \ENDIF
    \STATE Detecting duplicated vertex $\mathbf{Dv}$ and unreferenced vertex $\mathbf{Uv}$
    \ENDWHILE
\end{algorithmic}
\end{algorithm}
\subsection{Preprocessing and keypoint selection}
Keypoint is used to build local area correspondence between reference and distorted meshes. Similar method has been utilized in point cloud full-reference metrics \cite{yang2020inferring}\cite{meynet2020pcqm}\cite{yangMPED}. Point clouds can easily use an extant or add a pseudo vertex/point as keypoint to realize precise local area matching, since the units of the point cloud are scattered points that do not have connections. However, static meshes have a fixed geometry structure with polygonal face as the unit. Merging a new pseudo vertex to a surface without affecting the geometry is a relatively complex problem. 

To simplify the above problem, we propose selecting keypoints based on mesh vertices.  
For typical mesh distortions, such as mesh decimation, that will reduce the number of vertices, resulting in the distorted meshes having sparser vertex distributions while having larger face size than the reference meshes. A vertex, which coordinates copied from a certain vertex belong to the distorted mesh, might help to find a closer neighbor from the reference mesh vertex cluster, minimizing the shifting of local area matching. {Concurrently, sometimes the distorted meshes can have more vertices than the reference mesh, such as the SOTA compression tools, proposed by MPEG WG7 and AoMedia VVM, introduce subdivision in the decoding procedure expecting to reconstruct meshes with better quality. It reveals that simply using the extant mesh vertices as keypoints to construct local area correspondences is not enough.}

\begin{figure}[h]
	\centering
    \includegraphics[width=1.0\linewidth]{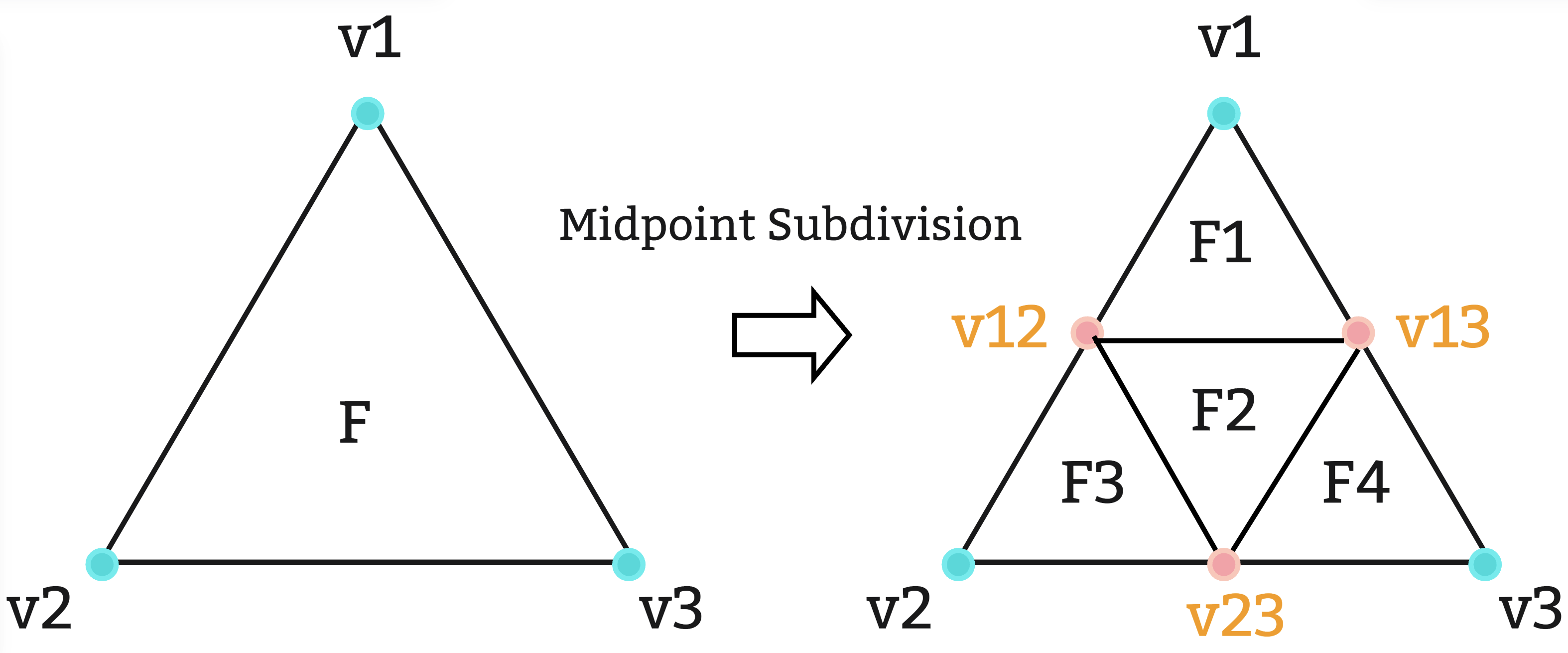}%
	\caption{Example of midpoint subdivision.}
	\label{fig:midsub}
\end{figure}

{Therefore, to generate better keypoint candidates, a two-step preprocessing using midpoint subdivision is first performed to equal the number of faces and vertices between the reference and distorted meshes. Specifically, given a single mesh triangle face, midpoint subdivision generates four new sub-faces with additional three new vertices, as shown in Fig. \ref{fig:midsub}. The mesh that has more faces ($\rm Na$) are selected as anchor mesh, then we calculate the times of performing midpoint subdivision for all the faces of the mesh that has fewer faces ($\rm Ns$) with the following rule:
\begin{align}
  \rm  \ max \ \mathbf{t}, s.t. \  4^\mathbf{t} \leq \frac{Na}{Ns}.
\end{align}
After performing $\mathbf{t}$ times' midpoint subdivision for all faces, a new mesh with $\rm 4^{\mathbf{t}}Ns$ faces is generated, which is the first step of preprocessing. 
}

{Then, a partial midpoint subdivision is applied for the case $\rm Na -  4^{\mathbf{t}}Ns > 0$, which is the second step of preprocessing. We select the $\mathbf{p}$ largest faces that $\mathbf{p}$ satisfies
\begin{align}
  \rm  min \ \mathbf{p}, s.t. \  3\mathbf{p} + 4^{\mathbf{t}}Ns \geq Na.
\end{align}
to perform midpoint point subdivision once again. 
}

{After two-step preprocessing, we convert the mesh that originally has fewer faces into a new mesh with $\rm 3\mathbf{p} + 4^{\mathbf{t}}Ns$ faces, resulting in two meshes that have the same or very close number of faces without introducing any visual distortion. }

Then, we use sampling to select vertices from reference mesh as keypoints, formulated as 
\begin{align}
    \mathcal{K} = \phi [\M_{\V r}]_{kn} \in \R^{kn\times3}.
\end{align}
$\phi[*]_{kn}$ samples $kn$ points from the distorted mesh vertex matrix $\M_{\V r}$, generating the keypoint set $\mathcal{K}$.  

\subsection{1-hop geodesic patch construction}
\begin{figure}[h]
	\centering
    \includegraphics[width=1.0\linewidth]{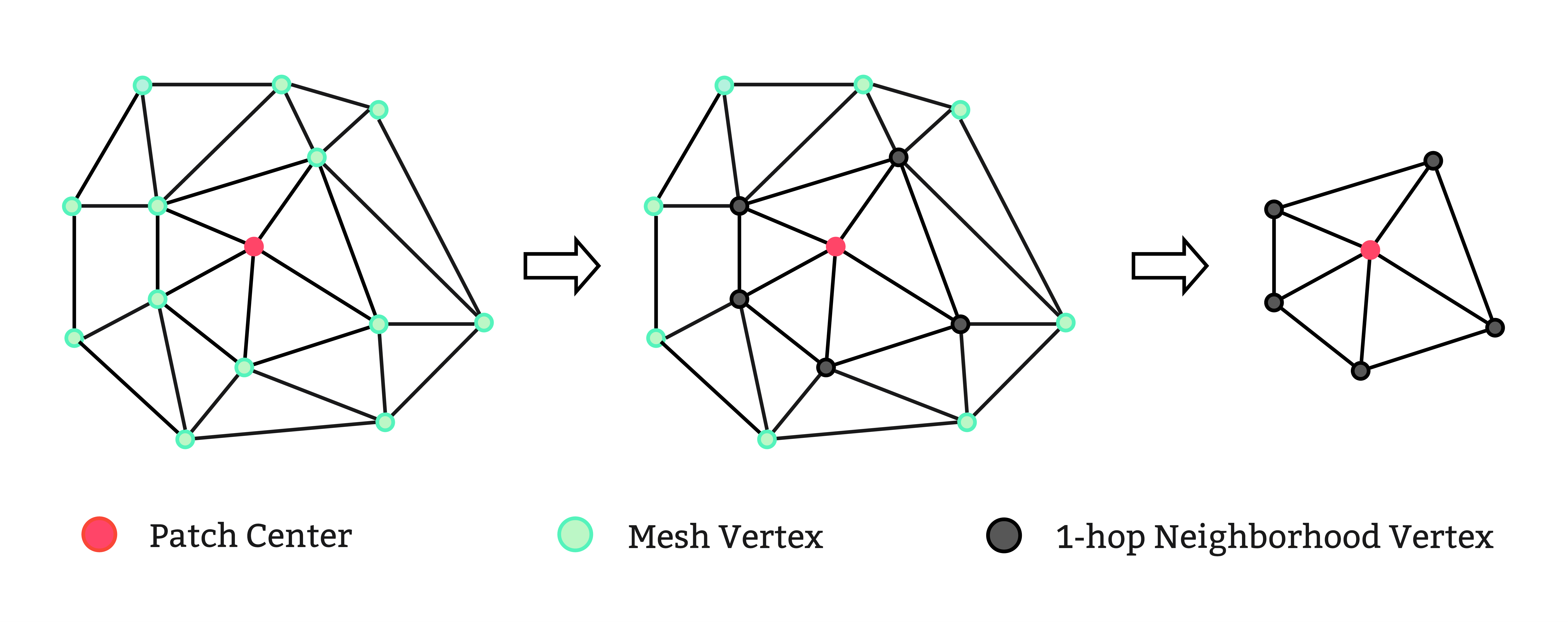}%
	\caption{Example of 1-hop Geodesic patch construction.}
	\label{fig:1-hop}
\end{figure}

The geometry information of static mesh is a natural 3D graph with vertices as graph nodes and faces to indicate graph edges. In GSP, if there is an edge between two vertices, then the two vertices are 1-hop neighbors to each other \cite{hu2021graph}. 1-hop geodesic patch consists of faces with 1-hop neighbors as vertices, exhibiting a mesh local geometry attribute.
Given the cleaned reference and distorted meshes, $\mathbf{M}_{r}$ and $\mathbf{M}_{d}$, for $k_{i}\in\mathcal{K}$, KNN search is first used to find $k_{i}$'s nearest neighbor in the reference and distorted meshes, $k_{ri}\in{\M_{\V r}}$ and $k_{di}\in{\M_{\V d}}$. Regarding the index of $k_{ri}$ and $k_{di}$ in ${\M_{\F r}}$ and ${\M_{\F d}}$, the 1-hop neighbors of $k_{ri}$ and $k_{di}$ can be easily extracted that appropriate for constructing 1-hop geodesic patch, i.e.,
\begin{equation}
\begin{aligned}
    \mathcal{V}_{(v_{r}, k_{ri})} & = \{v_{r}\} \subset \M_{\V r}, (v_{r}, k_{ri})\in \M_{\F r}, \\ 
    \mathcal{V}_{(v_{d}, k_{di})} & = \{v_{d}\} \subset \M_{\V d}, (v_{d}, k_{di})\in \M_{\F d}.
\end{aligned}
\end{equation}
$\mathcal{V}_{(v_{r}, k_{ri})}$ and $\mathcal{V}_{(v_{d}, k_{di})}$ represent the 1-hop neighbors of $k_{ri}$ and $k_{di}$ in reference and distorted meshes. $(v,k)\in \M_{\F}$ indicates vertices $v$ and $k$ form an edge of one or more faces collected by $\M_{\F}$. Fig. \ref{fig:1-hop} illustrates the toy example of 1-hop Geodesic patch construction.

 Two toy examples of 1-hop geodesic patches are illustrated in Fig. \ref{fig:flowchart}. Reference and distorted 1-hop geodesic patches centered with $k_{ri}$ and $k_{di}$ are recorded that $\mathbf{G}_{k_{ri}, \mathbf{M}_{r}}$ and $\mathbf{G}_{k_{di}, \mathbf{M}_{d}}$, with $\mathcal{V}_{(v_{r}, k_{ri})}$ and $\mathcal{V}_{(v_{d}, k_{di})}$ to construct the faces $\mathcal{F}_{\mathcal{V}_{(v_{r}, k_{ri})}}$ and $\mathcal{F}_{\mathcal{V}_{(v_{d}, k_{di})}}$, i.e., $\mathbf{G}_{k_{ri}, \mathbf{M}_{r}} = \mathcal{V}_{(v_{r}, k_{ri})} \oplus \mathcal{F}_{\mathcal{V}_{(v_{r}, k_{ri})}} $ and $\mathbf{G}_{k_{di}, \mathbf{M}_{d}} = \mathcal{V}_{(v_{d}, k_{di})} \oplus \mathcal{F}_{\mathcal{V}_{(v_{d}, k_{di})}} $, where $\mathbf{G}_{k,\mathbf{M}}$ means Geodesic patch with $k$ as patch center and faces from mesh $\mathbf{M}$, $a = b\oplus c$ means $b$ and $c$ are two components of $a$.

\subsection{Patch cropping}
Patch cropping aims to crop 1-hop geodesic patches to a predefined size, facilitating the extraction of effective features. 

There are two reasons to perform patch cropping. First, the reference and distorted 1-hop geodesic patches may cover areas of different sizes. For example, after mesh decimation, the decimated meshes usually have larger face sizes, resulting the distorted 1-hop geodesic patch including exclusive information that is out of coverage of the reference 1-hop geodesic patch.

The second reason is that the initial face size for different source meshes or different areas of the same source meshes might also be different, while structure features are generally sensitive to the size of the feature extraction unit. A properer feature extraction unit is prominent for feature purification.  Referring to the study on neuroscience and quality assessment \cite{simoncelli2001natural,yamins2016using,thibos1989image, wang2004ssim}, the perceptual visual signal can be simulated using a low-pass filter to aggregate information from the retina, indicating that the processing unit of the human visual system is limited in an area centered at the viewpoint. Excessively large patch sizes can weaken the effectiveness of feature extraction, consequently resulting in a deviation of human perception prediction.

\begin{figure}[h]
	\centering
    \includegraphics[width=1.0\linewidth]{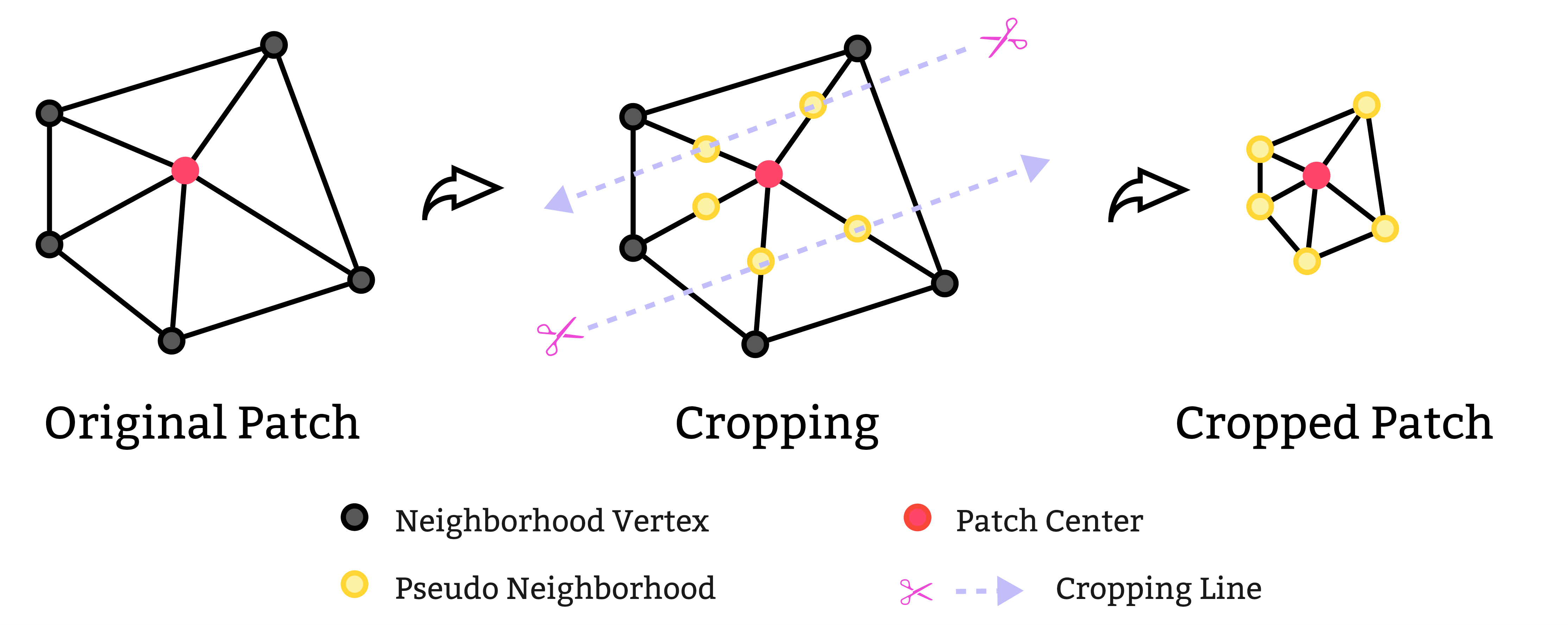}%
	\caption{Example of patch cropping.}
	\label{fig:cropping}
\end{figure}

Therefore, we propose a two-step patch cropping method shown in \textbf{Algorithm \ref{al:cropping}}: the cropping step 1 ensures that two 1-hop geodesic patches share close patch size, and the cropping step 2 ensures that the size of both the reference and distorted patches cannot be greater than a predefined threshold $\tau$. Specifically, the size of a patch is defined as the average distance between neighbors to the patch center. This cropping algorithm shrinks the 1-hop geodesic patch by generating pseudo 1-hop neighbors based on the original neighbors, resulting in similar polygons before and after cropping, which will not change the overall patch geometry characteristics. The position of patch cropping for the entire metric framework is highlighted by the red dashed lines in Fig. \ref{fig:flowchart}, a more vivid toy example of patch cropping is shown in Fig. \ref{fig:cropping}.

\begin{algorithm}[h]
\caption{Mesh Patch Cropping}
\label{al:cropping}
\begin{algorithmic}[1]
    \REQUIRE The initial reference and distorted 1-hop geodesic patches,$\mathbf{G}_{k_{ri}, \mathbf{M}_{r}}$ and $\mathbf{G}_{k_{di}, \mathbf{M}_{d}}$, a threshold $\tau$
    \ENSURE The cropped reference and distorted 1-hop geodesic patches,$\mathbf{G^{'}}_{k_{ri}, \mathbf{M}_{r}}$ and $\mathbf{G^{'}}_{k_{di}, \mathbf{M}_{d}}$
    \STATE Calculating the average Euclidean distance from the 1-hop neighbors to the patch center, i.e., $D_r = \frac{1}{|v_r|}\sum_{v_r^j\in \V_{v_r, k_{ri}}}||v_r^j-k_{ri}||_2$, $D_d = \frac{1}{|v_d|}\sum_{v_d^j\in \V_{v_d, k_{di}}}||v_d^j-k_{di}||_2$
    \STATE \textit{\textbf{Cropping step 1}}
    \IF{$(D_d/D_r = t) > 1$ }
    \STATE Cropping the distorted 1-hop geodesic patch $\mathbf{G}_{k_{di}, \mathbf{M}_{d}}$: generating the pseudo 1-hop neighbors  $\V_{v_d^{'}, k_{di}} = \{\frac{v_{d}^j-k_{di}}{t}+v_{d}^j\}$, generating the texture coordinates for $\V_{v_d^{'}, k_{di}}$  with the same method
    \ENDIF
    \IF{$(D_d/D_r = t) < 1$ }
    \STATE Cropping the reference 1-hop geodesic patch $\mathbf{G}_{k_{ri}, \mathbf{M}_{r}}$: generating the pseudo 1-hop neighbors  $\V_{v_r^{'}, k_{ri}} = \{\frac{v_{r}^j-k_{ri}}{t}+v_{r}^j\}$, generating the texture coordinates for $\V_{v_r^{'}, k_{ri}}$  with the same method
    \ENDIF
    \STATE \textit{\textbf{Cropping step 2}}
    \IF{$(D_r/\tau = l) >1$}
    \STATE Cropping the reference and distorted 1-hop geodesic patches: $\V_{v_r^{'}, k_{ri}} = \{\frac{v_{r}^j-k_{ri}}{\tau}+v_{r}^j\}$, $\V_{v_d^{'}, k_{di}} = \{\frac{v_{d}^j-k_{di}}{\tau}+v_{d}^j\}$, generating the texture coordinates for $\V_{v_r^{'}, k_{ri}}$ and $\V_{v_d^{'}, k_{di}}$, accordingly.
    \ENDIF
\end{algorithmic}
\end{algorithm}

\subsection{Patch texture mapping}
Geodesic patches only provide geometry information, i.e., vertex and face, it is not what the human eye actually perceives under actual viewing conditions, leading to the proposal of patch texture mapping.


Generally, a texture coordinate is a pair of floating numbers between 0 and 1 noted as $(u,v)$, given the pixel matrix of a texture map whose resolution is $L\times W$, the corresponding pixel index of $(u,v)$ is
\begin{equation}
\begin{aligned}
x = u\times L, y = (1-v) \times W.
\end{aligned}
\end{equation}
Each spatial face is textured by a 2D texture area consisting of a group of pixels \cite{tip-3-repre,tip-1-recon}. Using a triangle face as an example, the coordinates of the three vertex pixels are $(x_1, y_1)$, $(x_2, y_2)$, and $(x_3, y_3)$, circling a triangle on the texture map with a cluster of pixels contained, i.e.,
\begin{align}\label{eq:pixel_cluster}
    \mathcal{T}_{i, \mathbf{G}_{k, \mathbf{M}} } = \{p_{x,y}\}\subset \mathbf{Tri[(x_1, y_1),(x_2, y_2),(x_3, y_3)]}.
\end{align}
The face-wise pixel cluster, $\mathcal{T}_{i, \mathbf{G}_{k, \mathbf{M}} }$, circled by a 2D triangle $\mathbf{Tri[(x_1, y_1),(x_2, y_2),(x_3, y_3)]}$ transformed by the three vertex texture coordinates of the $ith$ face from the 1-hop geodesic patch $\mathbf{G}_{k, \mathbf{M}}$, is formulated in Eq. \eqref{eq:pixel_cluster}, with $\{p_{x,y}\}$ representing the pixels used to texture $ith$ face. For the whole 1-hop geodesic patch, we have a patch pixel cluster
\begin{equation}
\begin{aligned}
\mathcal{T}_{\mathbf{G}_{k, \mathbf{M}} } = & \{ \mathcal{T}_{i, \mathbf{G}_{k, \mathbf{M}} }\}, i \in |\mathcal{F}_{\mathcal{V}_{(v, k)}}|,
\end{aligned}
\end{equation}
$|\mathcal{F}_{\mathcal{V}_{(v, k)}}|$ represents the number of faces in the corresponding 1-hop geodesic patch.

After patch texture mapping, each 1-hop geodesic patch, i.e., $\mathbf{G}_{k, \mathbf{M}}$,  has a patch pixel cluster $\mathcal{T}_{\mathbf{G}_{k, \mathbf{M}} }$. To distinguish the patch that only has geometry information, we note the 1-hop geodesic patch with patch pixel cluster as the 1-hop textured geodesic patch, i.e., $\mathbf{TG}_{k, \mathbf{M}} = \mathbf{G}_{k, \mathbf{M}} \oplus \mathcal{T}_{\mathbf{G}_{k, \mathbf{M}} } $.
\subsection{Feature extraction}
Three types of feature are extracted from $\mathbf{TG}_{k, \mathbf{M}}$ to quantify the influence of distortion on human perception: patch color smoothness, patch discrete mean curvature, and patch pixel color average and variance.

\subsubsection{Patch color smoothness}
Patch color smoothness is inspired by the GSP theory \cite{shuman2013emerging}, which can reflect the signal values differences between neighboring vertices and has been widely used in image and point cloud denoising \cite{pcdenoising-zeng20193d, imgdenoising-bai2018graph}.  

Human perception is sensitive to high-frequency distortion, such as edge, contour, and texture gradient \cite{yang2019modeling}, motivating the graph representation of the 1-hop mesh geodesic patch to facilitate measurement of high-frequency features. An adjacency matrix, $\W$ , is first constructed as
\begin{equation}\label{adjancy matrix}
\mathbf{W}_{v_i,v_j}=
\begin{cases}
e^{-\frac{\|v_{i}-v_{j}\|^{2}_{2}}{2\sigma}}& \text{if } (v_{i}, v_{j})\in \mathcal{F}_{\mathcal{V}_{(v, k)}}, \\
0& \text{otherwise.}
\end{cases}
\end{equation}
Two vertices, $v_i$ and $v_j$, have an edge weight $\mathbf{W}_{v_i,v_j}$ if they were indicated with a definite connection relationship in the face matrix, i.e., $(v_{i}, v_{j})\in \mathcal{F}_{\mathcal{V}_{(v, k)}}$. Then the degree matrix, $\Dset$, used to measure the edge density of each vertex, is a diagonal matrix calculated by
$\Dset= {\sf diag}(d_{1},...,d_{n})\in \mathbb{R}^{n \times n}$, $d_{i}=\sum_{v_j} \mathbf{W}_{v_i,v_j}$, $n$ is the number of vertex in $\mathbf{G}_{k, \mathbf{M}}$. 

Both $\W$ and $\Dset$ lead to the graph Laplacian matrix,
\begin{align}
    \L=\Dset-\W,
\end{align}
which is a difference operand on graph. The normalized form of $\L$ is defined as $\L^{'}=\Dset^{-\frac{1}{2}}\L\Dset^{\frac{1}{2}}$ ~\cite{shuman2013emerging}.

For each vertex in $\mathbf{TG}_{k, \mathbf{M}}$, the three-channel RGB color information is provided with the pixel cluster. Considering YUV color space is suggested to be much closer related to human perception, a color space conversion is applied to generate graph signals as
\begin{align}\label{fc:rbg2yuv}
    f=[f_1, f_2, f_3] = {\rm [Y, U, V] = rgb2yuv([R, G, B]}),
\end{align}
$\rm rgb2yuv(*)$ realizes the color space conversion from $\rm [R, G, B]$ to $\rm [Y, U, V]$. A vertex might have multiple UV coordinates after mesh processings such as remeshing ~\cite{MPEG-MESH-cfp}, for this case, we calculate all the color with respect to different UVs and then average the results as the final vertex color.

The patch color smoothness for signal $f_i \in \mathbb{R}^{n\times1}$, $i=1,2,3$, is formulated as 
\begin{align}
    \mathbf{F}^{pcs}_{i} = \frac{f_i^{T}\L^{'}f_i}{|\mathcal{T}_{\mathbf{G}_{k, \mathbf{M}}}|}.
\end{align}
$|\mathcal{T}_{\mathbf{G}_{k, \mathbf{M}}}|$ is the number of pixels in $\mathcal{T}_{\mathbf{G}_{k, \mathbf{M}}}$, covered by $\mathbf{G}_{k, \mathbf{M}}$ on the texture map.

$f_i^{T}\L^{'}f_i$ is the Laplacian quadratic form of the graph in GSP, which is the integral of the gradient of the edge of the graph, appreciated for measuring the global smoothness of the signal ~\cite{shuman2013emerging}. Unlike the general representation of the graph in GSP, $\mathcal{T}_{\mathbf{G}_{k, \mathbf{M}}}$ is not a hollow graph with only vertices and edges. Regarding the definition of edge gradient in GSP, the number of effective pixels is used to normalize $f_i^{T}\L^{'}f_i$ and infer the patch color smoothness.

\subsubsection{Patch discrete mean curvature}
Patch discrete mean curvature is inspired by the intuition that the human visual system processes color and form separately \cite{rentzeperis2014distributed}, which can reflect the geometric mean curvature of the patch center. Patch color smoothness is mainly oriented to color features and only considers the influence of the Euclidean distance of the vertex while the spatial motion / orientation of the vertex is undervalued ~\cite{yang2020inferring}. To accurately infer the characteristics of the patch geometry, the discrete mean curvature of the patch is consequently calculated.

\begin{figure}[h]
	\centering
    \includegraphics[width=1.0\linewidth]{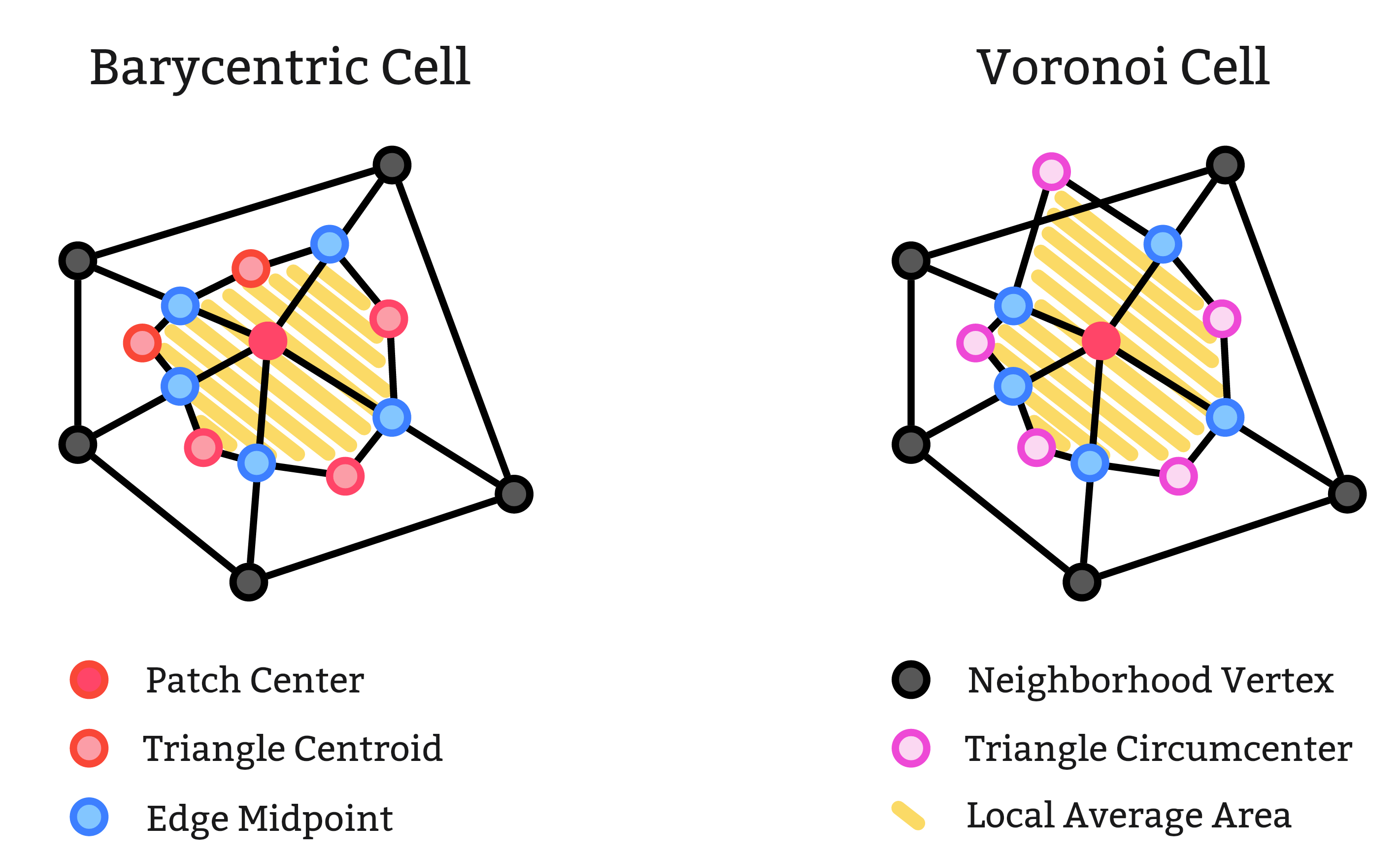}%
	\caption{Illustration of local averaging area.}
	\label{fig:local_area}
\end{figure}

To measure geometry deformation with respect to 3D space, the Laplace-Beltrami operand (LBO), which is the manifold generalization of the Laplace operand, is proposed \cite{lbo-do2016differential}. The Cotangent formula of LBO (Cot-LBO) is defined as
\begin{align}
    \L^{Cot}_i=\frac{1}{2A_{i}}\sum_{v_j}[cot(\alpha_{i,j})+cot(\beta_{i,j})](f(v_i)-f(v_j)),
\end{align}
$cot(\alpha_{i,j})$ and $cot(\beta_{i,j})$ are the cotangent values of two diagonals corresponding to edge $(v_i, v_j)$. $A_i$ is the local averaging area of $v_i$ which is an integral of the neighborhood of a point in the surface, such as the barycentric cell, the Voronoi cell and the mixed Voronoi cell, and is shown in Fig. \ref{fig:local_area}. $f$ is the vertex signal.

The essence of Cot-LBO is a vector in $v_i$, sharing the same magnitude with vertex mean curvature when $f(v) = v$\cite{lbodmc-meyer2003discrete}, it then leads to the discrete mean curvature $F_{i}^{dmc}$ of $v_i$:
\begin{align}
    &|\L^{Cot}_i| = |-2F_{i}^{dmc}\cdot N|, \\
    &|F_{i}^{dmc}| = \frac{|\L^{Cot}_i \cdot N|}{2}.
\end{align}
$N$ is the normalized normal vector at $v_i$, and we use the discrete mean curvature at the patch center to represent the patch discrete mean curvature, i.e., $F^{dmc} = F_{k}^{dmc}$.

\subsubsection{Patch pixel color average and variance}
Both patch color smoothness and discrete mean curvature are highly dependent on vertices, while texture inside the face is ignored. Therefore, the characteristics of textured faces are thus measured by patch pixel color average and variance, which can reflect the statistical features of face texture, i.e., mean and variance of luminance and chrominance.

Inspired by SSIM \cite{wang2004ssim}, the effective pixels of each face are analogized to the pixel window that is widely used in IQA. For the $ith$ face-wise pixel cluster, $\mathcal{T}_{i, \mathbf{G}_{k, \mathbf{M}} } $, of the patch pixel cluster $\mathcal{T}_{\mathbf{G}_{k, \mathbf{M}} } $, 
\begin{align}
   \mathcal{T}_{i, \mathbf{G}_{k, \mathbf{M}} } = \{p_{x,y}\} = \{[{\rm Y, U, V}]_{j}\}, j=1,2,...,m.
\end{align}
Each pixel provides a color information tuple $\rm [Y, U, V]_{j}$ after the same color space conversion as Eq. \eqref{fc:rbg2yuv}, and $m$ represents the number of pixels used to render the texture of $ith$ face. Channel-wise luminance and chrominance average and variance are derived as 
\begin{align}
   \overline{C_i} &= \frac{1}{M}\sum_{j}[C]_{j}, {C_i}^2 = \frac{1}{M}\sum_{j}([C]_{j}-\overline{C_i}), C\in\{{\rm Y, U, V}\}.
\end{align}
$[C]_j$ means value of color channel $C$ at the pixel index $j$. Patch pixel color average and variance are finally calculated via pooling all the face results, i.e.,
\begin{align}
  F_{1}^{pca} &= \frac{\sum_{i}s_i\overline{Y_i}}{\sum_{i}s_i}, F_{2}^{pca} = \frac{\sum_{i}s_i\overline{U_i}}{\sum_{i}s_i},F_{3}^{pca} = \frac{\sum_{i}s_i\overline{V_i}}{\sum_{i}s_i}, \\
    F_{1}^{pcv} &= \frac{\sum_{i}s_i Y^2}{\sum_{i}s_i}, F_{2}^{pcv} = \frac{\sum_{i} s_i U^2}{\sum_{i}s_i},F_{3}^{pcv} = \frac{\sum_{i}s_i V^2 }{\sum_{i}s_i}.
\end{align}
$s_i$ are the weighting factors between the faces, which are set according to the number of effective pixels of the face.

\subsection{Feature pooling}

\subsubsection{Patch similarity}
For each $\mathcal{T}_{\mathbf{G}_{k, \mathbf{M}} } $, we have patch color smoothness $F^{pcs}$, patch discrete mean curvature $F^{dmc}$, and patch pixel color average and variance $F^{pca}$ and $F^{pcv}$. $F^{pcs}$, $F^{pca}$ and $F^{pcv}$ are three triples corresponding to three color channels. Eq. \eqref{fc:sim} is proposed to measure the similarity of features extracted from patches constructed by keypoint $k$ from the reference and distorted meshes. 
\begin{align}\label{fc:sim}
\mathsf{SIM}^{F}_{k} =
\frac{|2F_{r}\cdot F_{d}|+T}{ (F_{r})^2+ (F_{d})^2+T}.
\end{align}
$T$ is a small non-zero constant to prevent numerical instability, $F\in\{F^{pcs}_i, F^{dmc}, F^{pca}_i, F^{pcv}_i\}$, $i\in \{1,2,3\}$. {For $F^{dmc}$, the reference and distorted mesh patches might have different directions, resulting in a positive and a negative values. Directly using Eq. \eqref{fc:sim} will incur bias. Therefore, we first convert all the values into positive ones. If different directions of curvature occur, we rescale $F^{dmc}_r = F^{dmc}_r + |F^{dmc}_r - F^{dmc}_d|$,  $F^{dmc}_d = F^{dmc}_d+ |F^{dmc}_r - F^{dmc}_d|$. }

A simple average pooling is first adopted to fuse feature similarities between multiple texture 1-hop geodesic patches, 

\begin{align}\label{fc:ave}
\mathsf{SIM}^{F}=\frac{1}{|\mathcal{K}|}\sum_{k\in\mathcal{K}}\mathsf{SIM}^{F}_{k}.
\end{align}
$\mathsf{SIM}^{F^{dmc}}$ can be calculated directly using Eq. \eqref{fc:ave_1}, e.g.,
\begin{align}\label{fc:ave_1}
\mathsf{SIM}^{F^{dmc}}=\frac{1}{|\mathcal{K}|}\sum_{k\in\mathcal{K}}\mathsf{SIM}^{F^{dmc}}_{k}.
\end{align}
Then weighting factors $\{\gamma_i\}_{i=1,2,3}$ are used to pooling triple features of  $F^{pcs}$, $F^{pca}$ and $F^{pcv}$ \cite{psnryuv-torlig2018novel}, e.g., 
\begin{align}\label{fc:color_pooling}
\mathsf{SIM}^{F}= \frac{1}{\gamma}\sum\nolimits_{i}\gamma_i\cdot \mathsf{SIM}^{F_i}, i\in \{1,2,3\},
\end{align}
where $\gamma_i$ is the pooling factor of $i$-channel reflecting the importance of individual color channels during visual perception, $\gamma=\sum\nolimits_{i}\gamma_i$.

In the end, we have the overall quality by averaging feature similarities:
\begin{equation}\label{fc:overall_pooling}
\mathrm{Q}=\frac{1}{4}[\mathsf{SIM}^{F^{pcs}} + \mathsf{SIM}^{F^{dmc}}+{\mathsf{SIM}^{F^{pca}}+\mathsf{SIM}^{F^{pcv}}}].
\end{equation}
\subsubsection{Patch similarity pooling}
Eq. \eqref{fc:ave} and \eqref{fc:ave_1} are used to average feature similarities between multiple texture 1-hop geodesic patches. Taking into account different degrees of matching and study related to visual attention mechanism \cite{min2020multimodal-at, min2016fixation-at2}, we divide patch pairs into three different types and calculate scores separately, followed by a pair-wise attention method to pooling the final objective score.

\begin{figure}[h]
	\centering
    \includegraphics[width=1.0\linewidth]{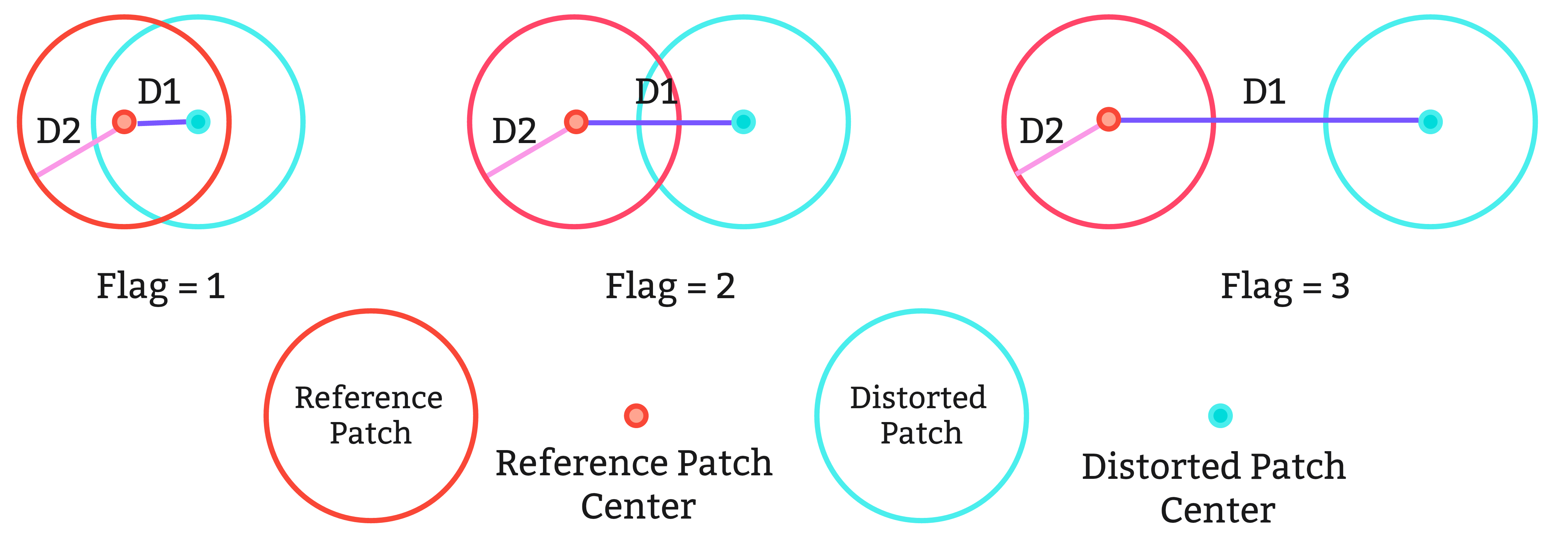}%
	\caption{Toy examples of patch classification.}
	\label{fig:patch_calss}
\end{figure}

$kn$ keypoints correspond to $kn$ pairs of 1-hop patches, the relationship between the distance of two patch centers (note as $\rm D1$) and the radius of the reference patch (note as $\rm D2$) are used to classify $kn$ pairs into three types by Eq. \eqref{fc: path_cla}:

\begin{equation}\label{fc: path_cla}
\mathbf{Flag}_{\{\mathbf{TG}_{k_{ri}, \mathbf{M}_{r}},\mathbf{TG}_{k_{di}, \mathbf{M}_{d}}\}}=
\begin{cases}
1& \text{if } \rm 0\ \leq D1 \leq D2, \\
2& \text{if } \rm D2\ < D1 \leq 2* D2, \\
3& \text{if } \rm D1 > 2* D2. \\
\end{cases}
\end{equation}
Toy examples of patch classification are shown in Fig. \ref{fig:patch_calss}. The classification rule can reflect the degree of patch shift taking the reference patch radius as reference. Considering quality metric expect to compare the exact local area difference for quality prediction, the priority of patch pair scores is decrease with the increase of patch shift.

Assume that the numbers of patch pairs of $\mathbf{Flag} = 1/2/3$ are $\mathbf{Flag}_{N1}$, $\mathbf{Flag}_{N2}$, $\mathbf{Flag}_{N3}$, respectively.
The patch similarity pooling as:
\begin{itemize}
    \item If patch pairs with $\mathbf{Flag} = 1/2$ do not exists, i.e., $\mathbf{Flag}_{N1} + \mathbf{Flag}_{N2} =0$, it means all the patch pairs are belong to $\mathbf{Flag} = 3$ and the mesh final quality score is calculated by Eq. \eqref{fc:ave} - \eqref{fc:overall_pooling} directly,  $\mathrm{S} = \mathrm{Q}_{\mathbf{Flag}_{N3}}$ where $\mathrm{S}$ represents the final objective score. The same is true for the cases where all the patch pairs belong to $\mathbf{Flag} = 1$ or $\mathbf{Flag} = 2$.
    \item If there exist patch pairs with $\mathbf{Flag} = 1/2$, i.e., $\mathbf{Flag}_{N1} + \mathbf{Flag}_{N2} > 0$, we use Eq. \eqref{fc:ave} - \eqref{fc:overall_pooling} to calculate two feature similarity scores corresponding to two classes respectively, i.e.,  $\mathrm{Q}_{\mathbf{Flag}_{N1}}$ and $\mathrm{Q}_{\mathbf{Flag}_{N2}}$. Patch pairs that $\mathbf{Flag} = 3$ are ignored. The final objective score is calculated as $\mathrm{S} = \frac{\mathrm{Q}_{\mathbf{Flag}_{N1}} + r \times \mathrm{Q}_{\mathbf{Flag}_{N2}}}{1 + r}$, where $r = \frac{\mathbf{Flag}_{N2}}{\mathbf{Flag}_{N1} + \mathbf{Flag}_{N2}}$.
\end{itemize}

\section{Experimental evaluations}\label{sec:exp}
This section first evaluates GeodesicPSIM and other SOTA metrics in the database introduced in section \ref{sec:TSMD}, exhibiting the superiority of the proposed GeodesicPSIM. Then, the robustness of GeodesicPSIM is explored with different hyperparameter settings.  The ablation study, presented in the last, demonstrates the importance of fusing three types of features and patch cropping.
\subsection{Metric parameters}
GeodesicPSIM has some hyperparameters and adaptive operands that need to be determined before application, leading to the following elaboration:

\begin{itemize}
	\item $ \phi [*]$, $kn$ in {\it keypoint selection}: $kn=5000$ and farthest point sampling (FPS) methods are used to extract keypoints.
        \item $\tau$ in {\it patch cropping}: $\tau = 0.5\times 10e^{-3} \times \mathrm{Bbox}$, $\mathrm{Bbox}=\max$($X_s$,$Y_s$,$Z_s$) with $X_s={X_{\max}-X_{\min}}$, $Y_s=Y_{\max}-Y_{\min}$ and $Z_s=Z_{\max}-Z_{\min}$ as respective bounding box scale of $x$-, $y-$, and $z$-axis of reference mesh. 
	\item  $\sigma$ in {\it patch color smoothness}: $\sigma$ is determined using the averaged Euclidean distance between vertices that have edges defined in $\mathcal{F}_{\mathcal{V}_{(v, k)}}$.
        \item $A_{i}$ in {\it patch discrete mean curvature}: Voronoi cell is used as a local averaging area to calculate Cot-LBO. 
	\item $T$, $\gamma_i$ in {\it feature pooling}: $T$ is set to $2.22\times e^{-16}$, which is the minimal floating number defined in MATLAB software. $[\gamma_{1},\gamma_{2}, \gamma_{3}] = [6,1,1]$ to reflect the different importance of various color components as proposed in $\rm PSNR_{yuv}$ \cite{psnryuv-torlig2018novel}, in which luminance is more sensitive to HVS. 
\end{itemize}

\subsection{SOTA metric introduction}
Three types of SOTA metrics are introduced, which will be used as comparative baselines to highlight the superiority of GeodesicPSIM: image-based, point-based, and video-based metrics. 
\subsubsection{Image-based metrics}
Image-based metrics, proposed by WG 7 \cite{MPEG-MESH-metric}, are based on images projected from 16 viewpoints resulting from the Fibonacci sphere lattice \cite{fibonacci}. Three metrics are tested based on the 16 images obtained: $\rm geo_{psnr}$, $\rm rgb_{psnr}$, and $\rm yuv_{psnr}$. $\rm geo_{psnr}$ calculates the depth information differences between the reference and distorted meshes. The depth information is captured as an image, which pixel values are normalized to 255, to get PSNR values comparable to the ones obtained using the next two metrics.  $\rm rgb_{psnr}$ and $\rm yuv_{psnr}$ calculate the differences of $\rm (R, G, B)$ and $\rm (Y, U, V)$ color channels between the reference and distorted images. {Besides, these metrics only consider effective projected pixels, which means the background content is ignored.}  In this paper, the resolution of the projected images is set as $1920\times1920$. A more detailed description of the metrics can be found in ~\cite{MPEG-MESH-metric}.

\subsubsection{Point-based metrics}
The point-based metrics are based on sampled point clouds from meshes. Seven point cloud quality metrics are evaluated: D1 \cite{D1-Mekuria2016Evaluation}, D2 \cite{tian2017geometric}, $\rm PSNR_{yuv}$ \cite{psnryuv-torlig2018novel},  PointSSIM \cite{pointssim-alexiou2020towards}, MPED \cite{yangMPED}, $\rm PCQM_{psnr}$ \cite{meynet2020pcqm}\cite{MPEG-MESH-metric}, and GraphSIM \cite{yang2020inferring}.  As ~\cite{MPEG-MESH-metric} reported that grid sampling has a stable behavior, it is used in this section with a grid resolution of 1024 to generate the colored point clouds.

\subsubsection{Video-based metric}
The video-based metrics are based on PVSs displayed during the subjective experiment \cite{yang2023tdmd}. The MSU Video Quality Measurement Tool ~\cite{MSU-antsiferova2022video} is used to compute the following IQA/VQA metrics: PSNR, SSIM ~\cite{wang2004ssim}, MSSIM ~\cite{wang2003multiscale}, 3SSIM ~\cite{li2009three}, VQM~\cite{vqm}, and VMAF ~\cite{vmaf-li2016toward}. PVSs have $1920\times1080$ resolution and a duration of 18 seconds.

\subsection{Performance of objective metrics}
To ensure the consistency between subjective scores (e.g., MOS) and objective predictions from various metrics, the objective predictions of different metrics are mapped to the same dynamic range following the recommendations suggested by the video quality experts group \cite{video2003final}. Three performance correlation indicators are employed to quantify the efficiency of the objective metrics: PLCC, SRCC, and RMSE. PLCC can demonstrate prediction accuracy, SRCC can reflect prediction monotonicity, and RMSE can show prediction consistency. More details can be found in~\cite{video2003final}.

\subsubsection{Correlation of objective metric}
We report the results of PLCC, SRCC, and RMSE on the whole database in columns "All" of Table \ref{TABLE-TSMDPer}. GeodesicPSIM reports the best performance, which is the one and only metric that PLCC and SRCC are higher than 0.8, followed by GraphSIM, $\rm PCQM_{psnr}$, and $\rm geo_{psnr}$.

\begin{table}[!ht]
    \centering
    	\caption{Metric performance on TSMD.} \label{TABLE-TSMDPer}
    \begin{scriptsize}
 \renewcommand{\arraystretch}{1.5}
	\setlength{\tabcolsep}{0.5mm}{
    \begin{tabular}{|c|c|c|c|c|c|c|c|c|c|c|c|c|}
    \hline
        ~ & ~ & \multicolumn{3}{c|}{All}& \multicolumn{6}{c|}{AOMedia Classes}  & \multicolumn{2}{c|}{Creation} \\ \hline
        Type & Metric & PLCC & SRCC & RMSE & A2 & B & D-1 & D-2 & E & F & DCC  & 3DS   \\ \hline
        \multicolumn{2}{|c|}{Number of Sample} & \multicolumn{3}{c|}{210} & 30 & 30 & 55 & 35 & 30 & 25 & 35  & 175   \\ \hline
        \multirow{3}{*}{$\bf A$} & $\rm geo_{psnr}$ & 0.73  & 0.73  & 0.80  & 0.92  & 0.90  & 0.72  & 0.71  & 0.87  & 0.74  & {\color{red}\textbf{0.93}}  & 0.73   \\ \cline{2-13}
         & $\rm rgb_{psnr}$ & 0.69  & 0.67  & 0.84  & 0.86  & 0.84  & 0.77  & 0.89  & 0.74  & 0.80  & 0.87  & 0.72   \\ \cline{2-13}
         & $\rm yuv_{psnr}$ & 0.68  & 0.68  & 0.85  & 0.85  & 0.83  & 0.76  & 0.88  & 0.76  & 0.80  & 0.86  & 0.72   \\ \hline
        \multirow{7}{*}{$\bf B$} & D1 & 0.72  & 0.65  & 0.80  & 0.93  & {\color{red}\textbf{0.95}}  & 0.73  & 0.68  & 0.65  & 0.77  & 0.92  & 0.71   \\ \cline{2-13}
         & D2 & 0.54  & 0.47  & 0.98  & 0.84  & 0.93  & 0.69  & 0.52  & 0.48  & 0.44  & 0.85  & 0.54   \\ \cline{2-13}
         & $\rm PSNR_{yuv}$ & 0.60  & 0.60  & 0.93  & 0.69  & 0.79  & 0.66  & 0.85  & 0.63  & 0.79  & 0.69  & 0.64   \\ \cline{2-13}
         & MPED & 0.48  & 0.50  & 1.01  & 0.56  & 0.39  & 0.44  & 0.50  & 0.66  & 0.75  & 0.52  & 0.48   \\ \cline{2-13}
         & PointSSIM & 0.48  & 0.41  & 1.02  & 0.85  & 0.52  & 0.56  & 0.72  & 0.59  & 0.43  & 0.75  & 0.46   \\ \cline{2-13}
         & $\rm PCQM_{psnr}$ & 0.76  & 0.75  & 0.76  & 0.85  & 0.85  & 0.81  & {\color{red}\textbf{0.91}}  & 0.76  & 0.89  & 0.82  & 0.78   \\ \cline{2-13}
         & GraphSIM & 0.79  & 0.79  & 0.70  & 0.90  & 0.85  & {\color{red}\textbf{0.86}}  & 0.77  & 0.77  & {\color{red}\textbf{0.96}}  & 0.91  & 0.78   \\ \hline
        \multirow{6}{*}{$\bf C$} & PSNR & 0.53  & 0.54  & 0.99&  {\color{red}\textbf{0.95}}  & 0.55  & 0.66  & 0.65  & 0.80  & 0.44  & 0.86  & 0.55   \\ \cline{2-13}
         & SSIM & 0.51  & 0.67  & 0.85  & 0.78  & 0.53  & 0.64  & 0.63  & 0.74  & 0.52  & 0.74  & 0.48   \\ \cline{2-13}
         & MSSIM & 0.66 & 0.65  & 0.87  & 0.92  & 0.51  & 0.77  & 0.74  & 0.89  & 0.54  & 0.86  & 0.64   \\ \cline{2-13}
         & 3SSIM & 0.68 & 0.67  & 0.85  & 0.94  & 0.52  & 0.83  & 0.73  & 0.93  & 0.52  & 0.91  & 0.69   \\ \cline{2-13}
         & VQM & 0.67 & 0.67  & 0.86  & 0.89  & 0.66  & 0.74  & 0.74  & 0.89  & 0.51  & 0.87  & 0.69   \\ \cline{2-13}
         & VMAF & 0.70  & 0.51  & 1.00  & 0.94  & 0.63  & 0.79  & 0.76  & {\color{red}\textbf{0.91}}  & 0.61  & 0.87  & 0.69   \\ \hline
        \multicolumn{2}{|c|}{GeodesicPSIM}  & {\color{red}\textbf{0.81}}  & {\color{red}\textbf{0.81}}  & {\color{red}\textbf{0.69}}  & 0.87  & 0.80  & 0.83  & 0.86  & 0.85  & 0.83 & 0.88  & {\color{red}\textbf{0.79}} \\ \hline
    \end{tabular}}
	\end{scriptsize}
\end{table}

\begin{figure}[h]
	\centering
    \includegraphics[width=1.0\linewidth]{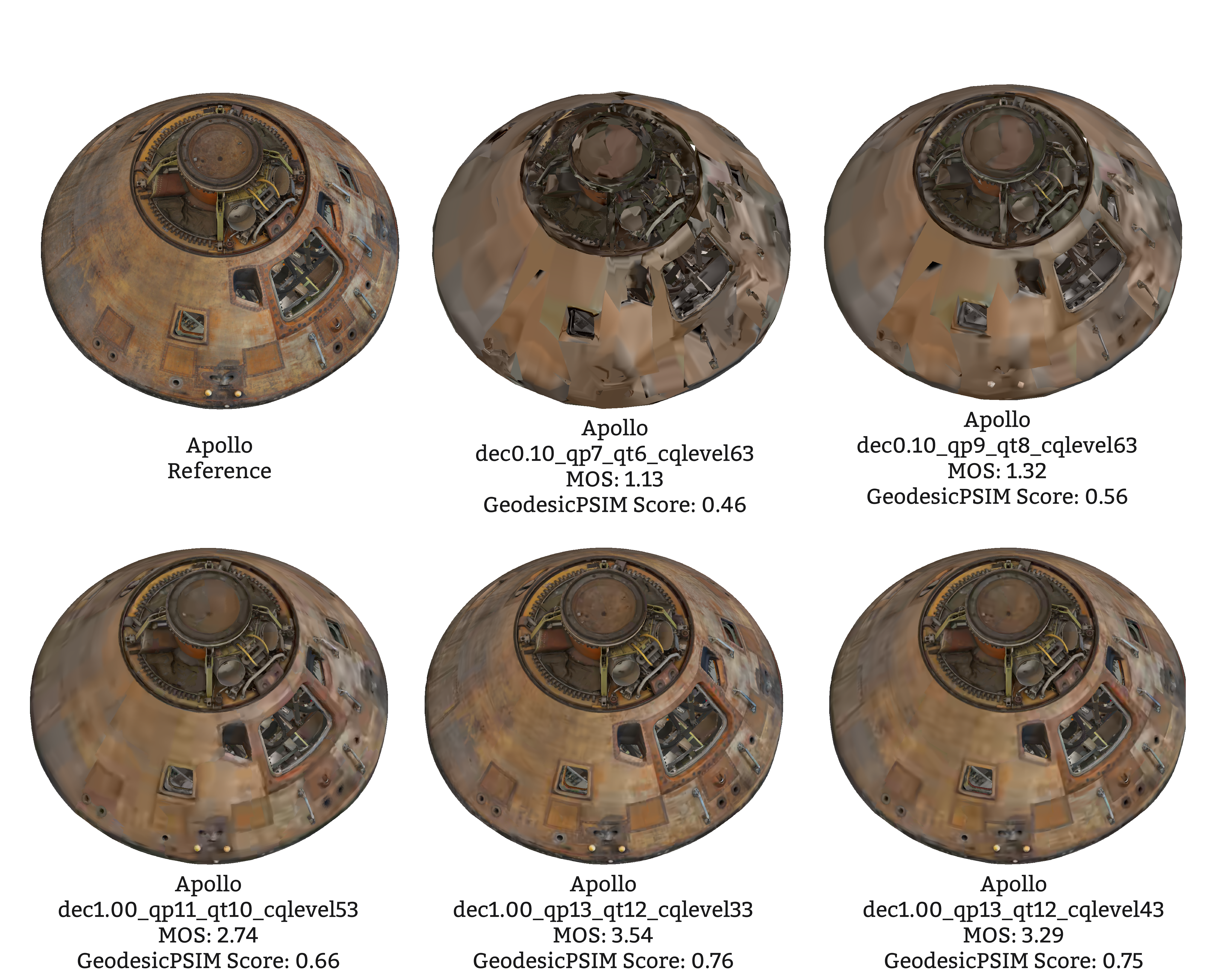}%
	\caption{Illustration of MOSs vs. GeodesicPSIM scores.}
	\label{fig:MosObj}
\end{figure}

\begin{figure}[pt]
	\centering
	\subfigure{		\includegraphics[width=0.48\linewidth]{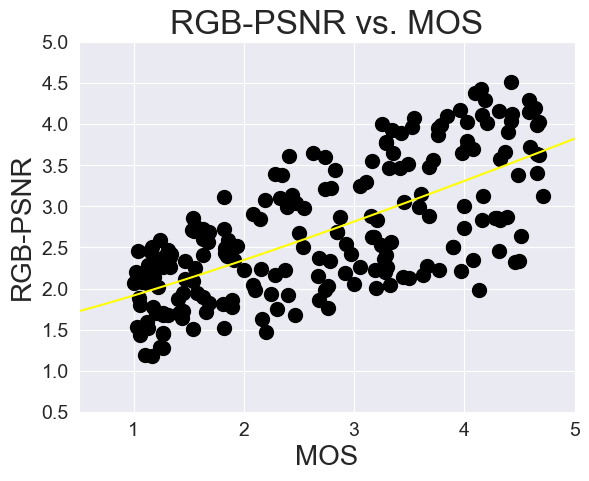}			}%
	\subfigure{		\includegraphics[width=0.48\linewidth]{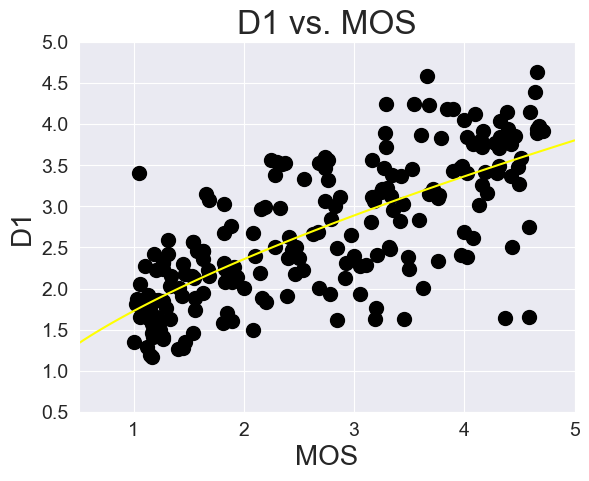}			}\\
	\subfigure{		\includegraphics[width=0.46\linewidth]{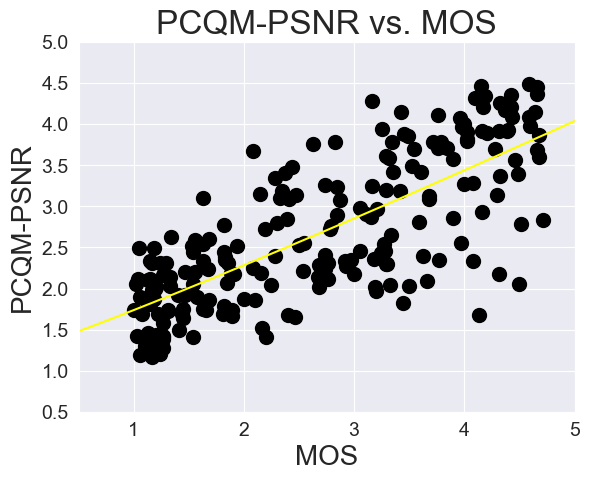}			} 
	\subfigure{		\includegraphics[width=0.46\linewidth]{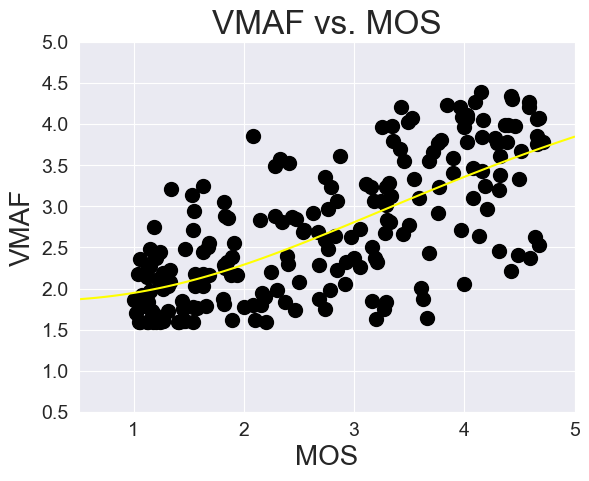}			}\\
 	\subfigure{		\includegraphics[width=0.46\linewidth]{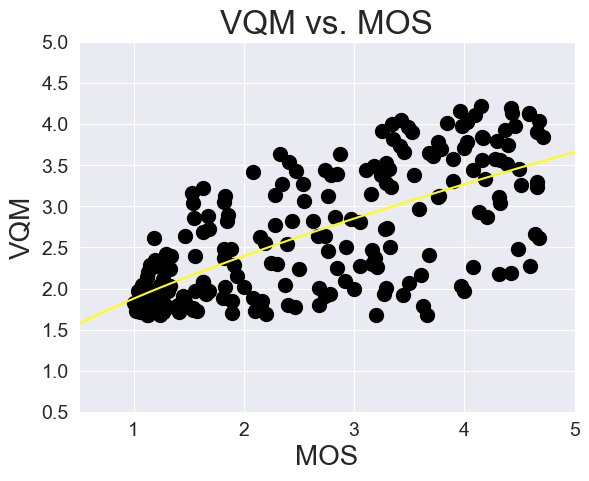}			}
 	\subfigure{		\includegraphics[width=0.46\linewidth]{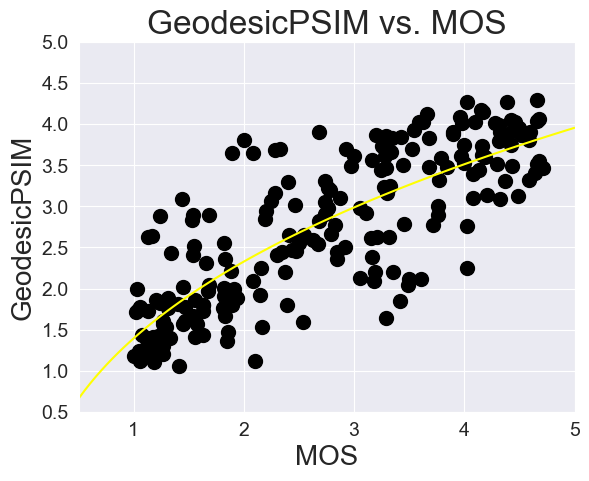}			}
	\caption{Scatter plots of metrics.}
	\label{fig:scatter_plots}
\end{figure}

Fig. \ref{fig:scatter_plots} shows the scatter plots of $\rm rgb_{psnr}$, D1, $\rm PCQM_{psnr}$, VMAF, VQM, and GeodesicPSIM. Points away from the best-fit logistic regression curve, colored in yellow, illustrating the bad prediction cases for each metric: GeodesicPSIM has the best scatter plot, with the yellow curve close to the perfect case ``$y=x$''. Fig. \ref{fig:MosObj} illustrates some samples with MOSs and GeodesicPSIM scores, also revealing the accuracy of the proposed metric. 
D1 reports several failed predictions with extreme value differences with MOSs, such as D1 = 3.5 vs. MOS = 1.0, D1 = 1.5 vs. MOS = 4.5, etc. VMAF tends to score lower for high-quality meshes. $\rm rgb_{psnr}$ has less outliers than $\rm PCQM_{psnr}$ but reports poorer quantified correlations, which seems contradictory, resulting in the analysis by content category exhibited in the next section.

\subsubsection{Analysis by category of content}
For a more in-depth analysis of the metric performance, the whole database is split into subclasses, sharing close characteristics or creation processes for each, with PLCC reported in columns "AOMedia Classes" and "Creation". Six different subclasses are defined in AOMedia VVM CfP \cite{VVM-cfp}. A2 collects game characters obtained by digital content creation (DCC), B is a collection of avatars captured as 3D scans (3DS) of humans, D consists of professional 3DS (D-1 with only triangular faces and D-2 with any polygonal faces), E contains non-professional 3DS of objects, and F professional 3DS from outdoor and indoor scenes. DCC and 3DS, which indicate the creation process of the meshes, also used as index to split database, the results are consequently reported in columns "Creation".

It is observed that the metrics that have the best "All" performance tend to present close results among different classes. Almost all the correlations of GeodesicPSIM are higher than 0.80 except 3DS is 0.79 and it demonstrates the best results on class 3DS. $\rm rgb_{psnr}$ shows close performance to $\rm yuv_{psnr}$ regardless of subclasses, indicating that the influence of the color space is limited on the pixel-wise metrics. D1 and D2 report the highest correlation on class B, revealing that they may be better suited to evaluate human figure meshes.  
$\rm PCQM_{psnr}$ reports better correlation than $\rm rgb_{psnr}$ in classes F and 3DS ($\rm \Delta PLCC > 0.05$), giving the reason that $\rm PCQM_{psnr}$ has a higher "All". Some video-based metrics show impressive results in classes A2 and E, with correlations above 0.90, but fail utterly in classes B and F. The score magnitude of video-based metrics is easily influenced by the proportion of background content in each video frame, which is not robust among different types of meshes, consequently resulting in performance gap for different classes. For instance, scores for indoor meshes are generally lower than scores for other meshes. The proportion of background content, static and artifact-free, contained in the frame significantly affects the range of quality scores. The frame for indoor meshes has no such background because the inside of the building covers the full frame. It consequently has more distorted information counted by the metrics, leading to a lower quality score.

\subsection{Robustness of GeodesicPSIM}
The robustness of GeodesicPSIM is evaluated in terms of the number of keypoint, the source of the keypoint, and the size of the 1-hop geodesic patch.
\subsubsection{The number of keypoint}
\begin{figure}[pt]
	\centering
	\subfigure[]{		\includegraphics[width=1\linewidth]{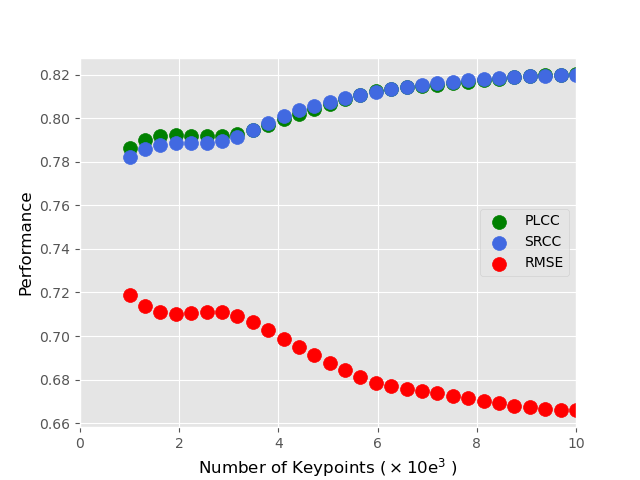}			} \\
	\subfigure[]{		\includegraphics[width=1\linewidth]{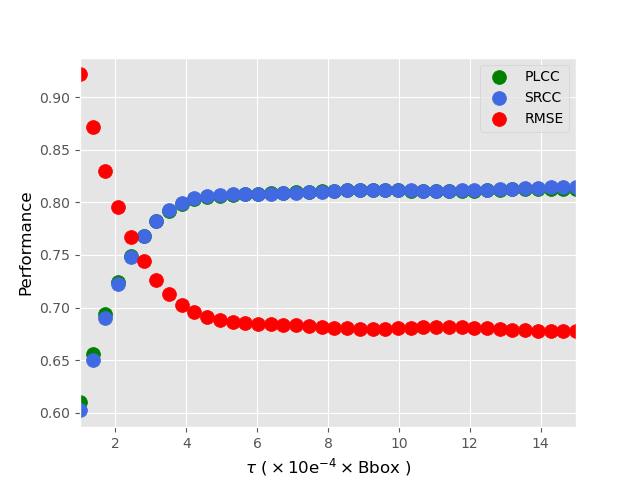}			} 
	\caption{(a): Metric performance with different number of keypoints; (b): Metric performance with different cropping thresholds.}
	\label{fig:plot}
\end{figure}

Keypoints play a vital role in GeodesicPSIM, maintaining the subsequent 1-hop geodesic patch construction and feature extraction. The performance of using a different number of keypoints to calculate GeodesicPSIM is reported in Fig. \ref{fig:plot} (a). The correlations gradually increase up to around 0.82 for PLCC and SRCC, 0.65 for RMSE, as the number of keypoints reaches 10000, and subsequently remain relatively stable, which demonstrates the model generalization with enough keypoints collected.

\subsubsection{The source of keypoint}
The keypoints are derived from the reference meshes after preprocessing with midpoint subdivision in section \ref{sec:exp}. To verify whether the source of the keypoint will significantly influence the performance of the metric, the results of using FPS to sample 5000 keypoints from different sources are reported in Table \ref{TABLE-pointSource}. It is observed that the metric shows better performance when the keypoints are sampled from the distorted meshes. We think the reason is that keypoints from the distorted mesh can better cover the distorted area, whereas keypoints from the reference meshes might ignore some local areas that have severe distortion.

\begin{table}[!ht]
\caption{Metric performance with different keypoint sources.} \label{TABLE-pointSource}
    \centering
        \begin{scriptsize}
  \renewcommand{\arraystretch}{1.5}
	\setlength{\tabcolsep}{0.7mm}{
    \begin{tabular}{|c|c|c|c|}
    \hline
         &  \multicolumn{3}{c|}{Correlation} \\ \hline
         Keypoint Source  & PLCC & SRCC & RMSE \\ \hline
        Reference  & 0.81	&0.81	&0.69 \\ \hline
        Distortion & 0.83	&0.82	&0.65 \\  \hline
    \end{tabular}}
    \end{scriptsize}
\end{table}

\subsubsection{The size of 1-hop geodesic patch}\label{sec:SoNe}
1-hop geodesic patch is the basic unit for quality prediction, which is sliced from the mesh surface. Given $\tau$ to control patch size, GeodesicPSIM fuses feature similarities on certain scales to infer quality scores. The performance of GeodesicPSIM  with respect to different $\tau$ is illustrated in Fig. \ref{fig:plot} (b). It reveals that the performance of the metric can be quickly improved by enlarging the patch size, reaching the highest correlation with PLCC and SRCC = 0.80 when $\tau$ around $\rm 4.5 \times 10e^{-4} \times Bbox$, and then maintaining stable. The curve implies that a proper patch size is extremely important for effective feature extraction.


\subsection{Ablation study}
This section examines GeodesicPSIM by reassembling its modules to report its efficiency.
\subsubsection{Feature extraction }
GeodesicPSIM pools three types of features, that is, patch color smoothness $F^{pcs}$, patch discrete mean curvature $F^{dmc}$, and patch pixel color average and variance $F^{pca}$ and $F^{pcv}$, to infer mesh quality. To validate feature effectiveness, the performance of the individual features is reported in Table \ref{TABLE-featurecombine}. $F^{dmc}$ exhibits the best results, followed by $F^{pcv}$, and $F^{pca}$ and $F^{pcs}$. Each feature contributes to the overall performance of GeodesicPSIM.

\begin{table}[!ht]
\caption{Feature effectiveness of GeodesicPSIM.} \label{TABLE-featurecombine}
    \centering
            \begin{scriptsize}
  \renewcommand{\arraystretch}{1.5}
	\setlength{\tabcolsep}{0.7mm}{
    \begin{tabular}{|c|c|c|c|}
    \hline
        Features Combination & PLCC & SRCC & RMSE \\ \hline
        $F^{pcs}$ & 0.39&	0.35&	1.07  \\ \hline
        $F^{dmc}$ & 0.79&	0.79&	0.72  \\ \hline
        $F^{pca}$  &0.52&	0.51&	0.99  \\ \hline
        $F^{pcv}$ &0.53&	0.53&	0.99  \\ \hline
        GeodesicPSIM & 0.81 & 0.81 & 0.69 \\ \hline
    \end{tabular}}
    \end{scriptsize}
\end{table}

\subsubsection{Key modules} \label{sec:abl-crop}
To reveal the effectiveness of two key modules, i.e., patch cropping and patch similarity pooling, we further investigate the contribution of different components and report the results in Table \ref{tab:ablation_module}. We see that each key module contributes to the overall performance, which verifies the effectiveness of the proposed technologies.

\begin{table}[h]
  \centering
  \caption{Ablation study for the key modules. `\Checkmark' or `\XSolidBrush' means the setting is preserved or discarded.}
    \begin{tabular}{cc|ccc}
    \toprule
    \multicolumn{1}{l}{Patch cropping} & \multicolumn{1}{l}{Patch similarity pooling} & \multicolumn{1}{l}{PLCC} & \multicolumn{1}{l}{SROCC}  & \multicolumn{1}{l}{RMSE} \\
    \midrule
    \XSolidBrush & \Checkmark   &0.799	&0.802	&0.699
 \\
    \Checkmark     & \XSolidBrush        &0.799	&0.798	&0.700

    \\
    \Checkmark     & \Checkmark        &\bf{0.806}	&\bf{0.807}	&\bf{0.688}

  \\
    \bottomrule
    \end{tabular}%
  \label{tab:ablation_module}
   \vspace{-0.4cm}
\end{table}%
\section{Conclusions}
\label{sec:conclusion}
In this paper, we propose a new model-based static mesh objective quality metric, called GeodesicPSIM. GeodesicPSIM consists of seven steps: mesh cleaning, keypoint selection, 1-hop geodesic patch construction, patch cropping, patch texture mapping, feature extraction, and feature pooling. We have used a newly created and challenging database, TSMD, to validate the superiority of GeodesicPSIM, on which GeodesicPSIM reports PLCC, SRCC, and RMSE at 0.81, 0.81, and 0.69, ranking first among all metrics tested. The robustness of GeodesicPSIM is examined by checking the performance variation in terms of different high-parameter settings, and ablation studies have shown the effectiveness of the three types of feature proposed and the novel patch cropping module. 

There are many expected tracks for future study. First, quality metrics for image, video, and point clouds have reported PLCC and SRCC above 0.90 on some databases, inspiring us to continue to improve the performance of GeodesicPSIM. Second, considering GeodesicPSIM is a full-reference metric that requires the injection of reference meshes, an effective no-reference metric needs to be designed to deal with the cases where the reference meshes are not available.


\ifCLASSOPTIONcaptionsoff
  \newpage
\fi

\bibliography{sample-base}

\bibliographystyle{IEEEtran}
\begin{IEEEbiography}
[{\includegraphics[width=1in,height=1.25in,clip,keepaspectratio]{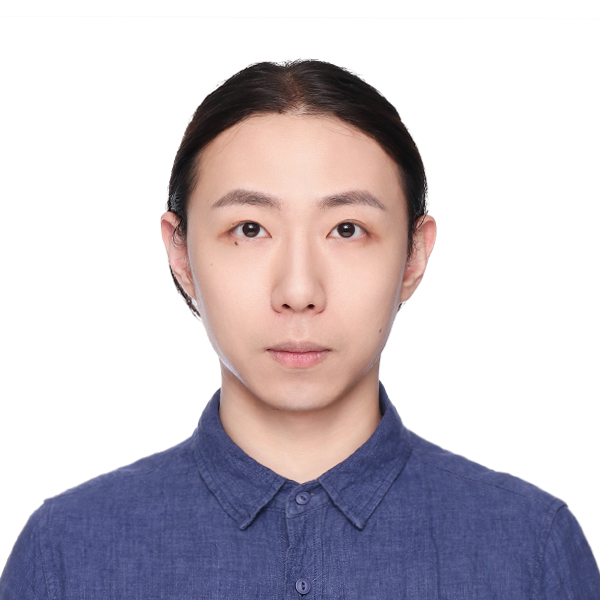}}]{Dr. Qi Yang} received the B.S. degree in communication engineering from Xidian University, Xi'an, China, in 2017, and Ph.D degree in information and communication engineering at Shanghai Jiao Tong University, Shanghai, China, 2022. Now, he is a researcher in Tencent MediaLab. He has published more than 20 conference and journal articles, including TPAMI, TVCG, TMM, ACM TOMM, CVPR, ACM MM, ICME, etc. He is also an active member in standard organizations, including MPEG, AOMedia, and AVS. His research interests include image quality assessment, 3D media compression and quality assessment.
\end{IEEEbiography}

\begin{IEEEbiography}
[{\includegraphics[width=1in,height=1.25in,clip,keepaspectratio]{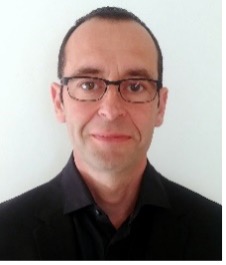}}]{Dr. Joel Jung} received the habilitation degree in electrical engineering from Sorbonne University, Paris, France, in 2019, and the Ph.D. degree in electrical engineering from the
University of Nice, France, in 2000. 

He joined Tencent Media Lab in 2020, and is currently engaged in activities centered around mesh compression and mesh quality assessment, focusing on two key aspects: 1- Actively contributing to Tencent's technical response to the AOMedia Call for Proposals on Static Polygonal Mesh Coding, 2- Playing a major role in designing efficient objective video quality metrics for meshes. As an active participant in MPEG standardization, he serves as the chair of the Immersive Video focus group within Advisory Group 5, to provide support to MPEG Working Groups on subjective and objective video quality issues. 
In previous roles, he was involved in video quality evaluation for gaming content and immersive video content, contributing to ITU-T Study Group 12 (Performance, QoS and QoE). He co-chaired the MPEG Immersive Video group for five years, overseeing the creation of standards for compressing multi-view content to facilitate 6DoF rendering. This involved expertise in multi-view, light-field representation formats, depth estimation, and the synthesis of intermediate viewpoints for seamless 6 degrees of freedom navigation. Additionally, I actively contributed to the standardization of MV-HEVC and 3D-HEVC, with ownership of the "Quadtree Limitation and Prediction" tool.

From 1996 to 2000, he was with the CNRS Laboratory, Sophia Antipolis, active in the improvement of video decoders based on the correction of compression and transmission artifacts. In 2000, he joined Philips Research France, Paris, as a Research Scientist in video coding, postprocessing, perceptual models, objective quality metrics, and low-power codecs. He worked at Orange Labs, France, from 2004 to 2020. He has contributed to the 2-D and 3-D video coding standard HEVC/3-D HEVC. 
and being chair of the immersive video focus group of MPEG Advisory Group 5 (AG5) on video quality assessment. In addition, he is involved in the standardization of immersive video codecs, as a co-chair of MPEG Immersive Video (MIV) group, dealing with coding, view synthesis, and depth estimation with six degrees of freedom.

\end{IEEEbiography}

\begin{IEEEbiography}
[{\includegraphics[width=1in,height=1.25in,clip,keepaspectratio]{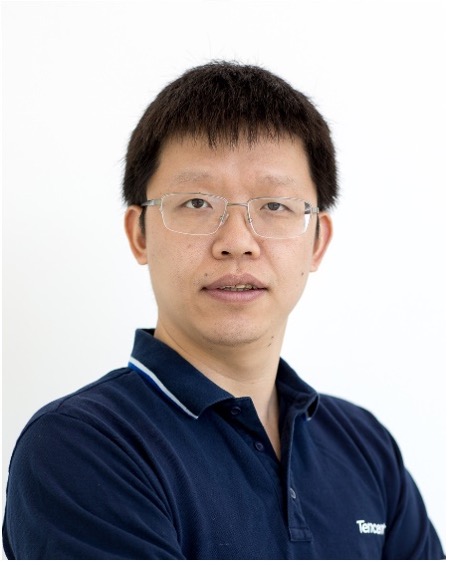}}]{Dr. Xiaozhong Xu} (Senior Member, IEEE) has been a Senior Principal Researcher and Senior Manager of Multimedia Standards at Tencent Media Lab, since 2017. He was with MediaTek as a Senior Staff Engineer and Department Manager of Multimedia Technology Development, from 2013 to 2017. Prior to that, he worked for Zenverge (acquired by NXP in 2014), a semiconductor company focusing on multi-channel video transcoding ASIC design, from 2011 to 2013. He also held technical positions at Thomson Corporate Research (now Technicolor) and Mitsubishi Electric Research Laboratories. His research interest lies in the general area of multimedia, including video, image and volumetric data coding, processing and transmission. He has been an active participant in various multimedia standardization activities for over fifteen years. Xiaozhong Xu received the B.S. and Ph.D. degrees from Tsinghua University, Beijing China in electronics engineering, and the MS degree from Polytechnic school of engineering, New York University, NY, USA, in electrical and computer engineering. He was a recipient of ISO\&IEC Excellence Award, Technology Lumiere Award, and Science and Technology Award from China Association for Standardization.
\end{IEEEbiography}

\begin{IEEEbiography}
[{\includegraphics[width=1in,height=1.25in,clip,keepaspectratio]{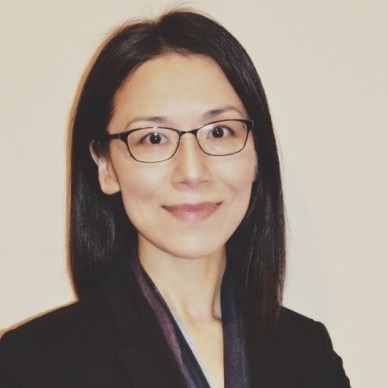}}]{Dr. Shan Liu} (Fellow, IEEE) received the B.Eng. degree in electronic engineering from Tsinghua University, the M.S. and Ph.D. degrees in electrical engineering from the University of Southern California, respectively. She is a Distinguished Scientist and General Manager at Tencent. She was formerly Director of Media Technology Division at MediaTek USA. She was also formerly with MERL and Sony, etc. She has been a long-time contributor to international standardization with many technical proposals adopted into various standards such as VVC, HEVC, OMAF, DASH, MMT and PCC, and served as a Project Editor of ISO/IEC | ITU-T H.266/VVC standard. She is a recipient of ISO\&IEC Excellence Award, Technology Lumiere Award, USC SIPI Distinguished Alumni Award, and two-time IEEE TCSVT Best AE Award. She currently serves as Associate Editor-in-Chief of IEEE Transactions on Circuits and Systems for Video Technology and Vice Chair of IEEE Data Compression Standards Committee. She also serves and has served on a few other Boards and Committees. She holds more than 600 granted US patents and has published more than 100 peer-reviewed papers and one book. Her interests include audio-visual, volumetric, immersive and emerging multimedia compression, intelligence, transport and systems.
\end{IEEEbiography}

\end{document}